\documentclass[10pt,twocolumn,letterpaper]{article}

\usepackage[pagenumbers]{cvpr}              %

\newcommand{\duster}[0]{{DUSt3R}}
\newcommand{\master}[0]{{MASt3R}}

\newcommand{\ours}[0]{{Pow3R}}
\newcommand{\embed}[0]{embed}
\newcommand{\inject}[1]{inject-$#1$}

\newcommand{\myparagraph}[1]{\vspace{0.5mm} \noindent \textbf{#1}}

\newcommand{\I}[1]{I^{#1}} %
\newcommand{\C}[1]{C^{#1}} %
\newcommand{\R}{\mathbb{R}} %
\newcommand{\D}[1]{D^{#1}} %
\newcommand{\X}[2]{X^{#1,#2}} %
\newcommand{\Xgt}[2]{\hat{X}^{#1,#2}} %
\newcommand{\z}{z}           %
\newcommand{\zgt}{\hat{z}}   %
\newcommand{\Z}{\text{norm}} %

\newcommand{\lreg}{\mathcal{L}^{\text{regr}}} %
\newcommand{\lconf}{\mathcal{L}^{\text{conf}}} %

\newcommand{\valid}{\mathcal{D}} %

\DeclareMathOperator*{\argmin}{arg\,min}
\newcommand{\un}[0]{\underline }

\definecolor{cvprblue}{rgb}{0.21,0.49,0.74}
\usepackage[pagebackref,breaklinks,colorlinks,allcolors=cvprblue]{hyperref}

\usepackage{bbding} %
\usepackage{multirow}
\usepackage{rotating}
\usepackage{units}

\def\paperID{932} %
\def\confName{CVPR}
\def\confYear{2025}

\title{\ours{}: Empowering Unconstrained 3D\\Reconstruction with Camera and Scene Priors}

\author{
Wonbong Jang\textsuperscript{*} \quad Philippe Weinzaepfel\textsuperscript{†} \quad Vincent Leroy\textsuperscript{†} \\
Lourdes Agapito\textsuperscript{*} \quad Jerome Revaud\textsuperscript{†} \\[0.1cm]
\textsuperscript{*}UCL \qquad \textsuperscript{†}Naver Labs Europe \\[-0.1cm]
{\tt\small \{ucabwja,l.agapito\}@ucl.ac.uk} \quad
{\tt\small firstname.lastname@naverlabs.com}
}

\begin{document}
\twocolumn[{%
\renewcommand\twocolumn[1][]{#1}%
\maketitle
\vspace{-0.8cm}
\begin{center}
    \captionsetup{type=figure}
    \includegraphics[width=\linewidth,trim=0 250 0 0,clip]{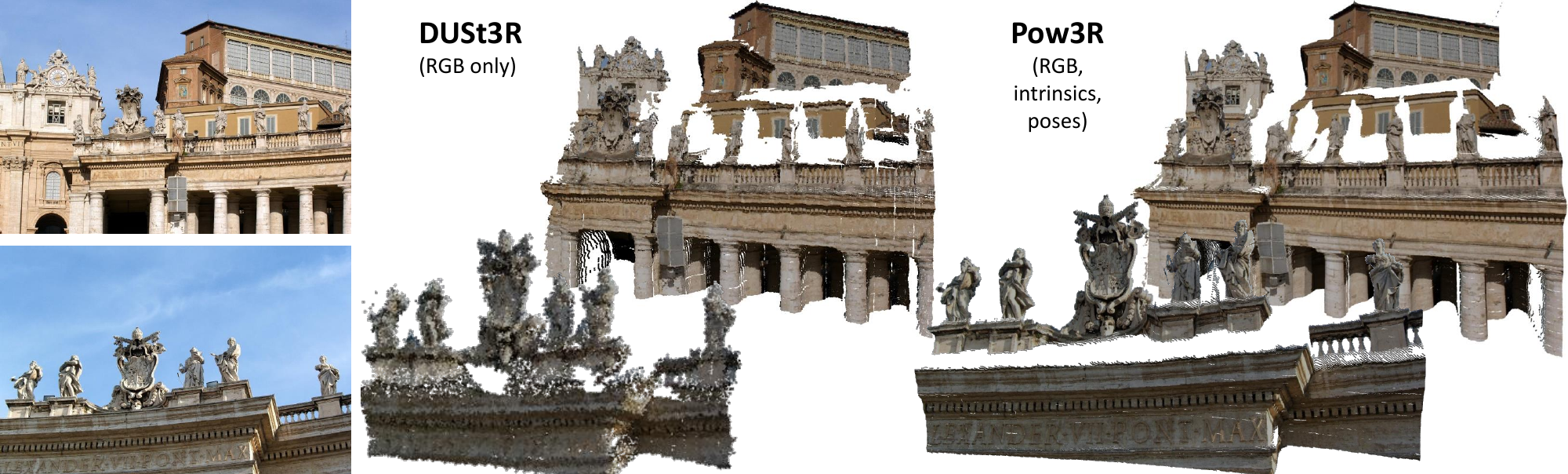}  \\[-0.3cm]
    \captionof{figure}{
    \textbf{Qualitative comparison} of raw 3D reconstructions between \duster{} and \ours{} for the image pair on the left. \duster{} cannot leverage auxiliary information potentially available at test time, resulting in noisy and low-resolution estimates, whereas \ours{} allows to inject such priors at test time in a versatile manner, enabling new capabilities such as high-resolution processing as a direct by-product. 
    }
    \vspace{-0.01cm}
    \label{fig:qualiCC}
\end{center}
}]

\begin{abstract}
\vspace{-0.5cm}

We present \ours{}, a novel large 3D vision regression model that is highly versatile in the input modalities it accepts. Unlike previous feed-forward models that lack any mechanism to exploit known camera or scene priors at test time, \ours{} incorporates any combination of auxiliary information such as intrinsics, relative pose, dense or sparse depth, alongside input images, within a single network. 
Building upon the recent \duster{} paradigm, a transformer-based architecture that leverages powerful pre-training, %
our lightweight and versatile conditioning acts as additional guidance for the network to predict more accurate estimates when auxiliary information is available. 
During training we feed the model with random subsets of modalities at each iteration, which enables the model to operate under different levels of known priors at test time.
This in turn opens up new capabilities, such as performing inference in native image resolution, or point-cloud completion.
Our experiments on 3D reconstruction, depth completion, multi-view depth prediction, multi-view stereo, and multi-view pose estimation tasks yield state-of-the-art results and confirm the effectiveness of \ours{} at exploiting all %
available information.
The project webpage is \href{https://europe.naverlabs.com/pow3r}{https://europe.naverlabs.com/pow3r}.

\vspace{-0.5cm}
\end{abstract}    
\section{Introduction}
\label{sec:intro}

Building general models for 3D perception that can unify different 3D vision tasks such as depth estimation, keypoint matching, dense reconstruction or camera pose prediction, remains a complex and exciting open challenge. 
Traditional Structure from Motion (SfM) approaches such as COLMAP~\cite{schonberger2016structure} %
do not apply pre-learned priors, so each scene must be optimized independently. 
On the other hand, most learning-based approaches focus on specific 3D vision tasks 
such as depth completion~\cite{completionformer2023, rho2022guideformer}, monocular and stereo depth estimation~\cite{,piccinelli2024unidepth}, pose estimation~\cite{wang2023posediffusion, raydiffusion} or novel view synthesis~\cite{mildenhall2021nerf, kerbl20233d}. While these dedicated approaches have been successful to some extent, this trend contrasts with other domains such as natural language processing where a single model like ChatGPT~\cite{openai2024chatgpt4}, trained with a vast collection of data, solves a range of different tasks without requiring re-training.
While some models like CroCo-v2, DINOv2 or Stable Diffusion~\cite{darcet2023dinov2, Rombach_2022_CVPR, weinzaepfel2023croco}, trained on large image datasets, have served as learned priors for other tasks, including 3D vision ones, they are still far from multi-purpose foundation models like ChatGPT is for NLP.

\duster{}~\cite{wang2024dust3r} marked a significant step forward in regression-based 3D foundation models. 
\duster{} performs 3D dense reconstruction from uncalibrated and unposed images, solving many downstream vision tasks at once  without the need for fine-tuning. Its success can be attributed to two main factors: (\emph{i}) its use of a pointmap representation, which provides flexibility over its output space and enables to regress camera parameters, poses and depth easily and (\emph{ii}) its use of a transformer architecture that exploits powerful pretraining combined with a vast amount of 3D training data. 
{Despite its huge success, \duster{}~\cite{wang2024dust3r} and its successor \master{}~\cite{leroy2024mast3r} are extremely limited in terms of their input space, \ie they only take RGB images.} 
Yet, in practice, many real-world applications \emph{do} provide extra modalities such that calibrated camera intrinsics, sparse or dense depth from RGB-D sensors or LIDAR, \etc.
This raises the question of how best to exploit additional input modalities, and whether this not just boosts performance but also enables new capabilities, such as obtaining the full depth from sparse point clouds of SfM or LiDAR.

In this paper, we introduce \ours{}, a new 3D feed-forward model that empowers DUSt3R with any subset of priors available at test time such as camera intrinsics, sparse or dense depth, or relative camera poses.
Each modality is injected into the network in a lightweight fashion. To allow the model to operate under different conditions at test time, random subsets of input modalities are fed at each training iteration.  
As a result, we obtain a single model that performs on par with DUSt3R when no prior information is available but outperforms it when it exists.
Our model also gains new capabilities as a by-product: for instance, the camera intrinsics input allow to process images whose principal point is far from the center, thus allowing to perform extreme cropping, \eg for performing sliding window inference. As a side contribution, our model directly outputs the pointmaps of the second image in its coordinate system, allowing faster relative pose estimation. To sum up, our contributions are:
\begin{itemize}
    \item \ours{} is a holistic 3D geometric vision model capable of taking any subset (including none) of camera intrinsics, pose and depthmaps with corresponding input images.
    \item Extensive experiments demonstrate an important boost in performance over \duster{} which cannot exploit priors, yielding state-of-the-art results in three benchmarks. %
    \item By predicting the same pointmaps in two different camera coordinate systems, we can achieve more accurate relative pose, orders of magnitude faster.
\end{itemize}

\section{Related work}
\label{sec:related}

\myparagraph{Structure-from-Motion and MVS.} 
SfM ~\cite{hartley2003multiple, crandall2012sfm, jiang2013global,cui2017hsfm, schonberger2016structure} builds 3D sparse maps by optimizing camera parameters using only RGB images. 
Traditional approaches consists of handcrafted pipelines to match keypoints ~\cite{harris1988combined, lowe2004distinctive, bay2006surf, rosten2006machine, barroso2019key}, form pixel correspondences across multiple images, determine geometric relations, and perform bundle adjustment to optimize camera locations.
Recently, SfM has incorporated learning-based approaches leveraging neural networks as components~\cite{yi2016lift, barath2023affineglue, chen2022aspanformer, lindenberger2023lightglue, sarlin2020superglue, sun2021loftr, tang2022quadtree, wang2023guiding}, but it remains brittle to outliers and challenging conditions.
Likewise, multi-view stereo (MVS) aims to reconstruct dense 3D surface through handcrafted pipelines based on triangulation from multiple viewpoints~\cite{furukawa2015multi, galliani2015massively, schonberger2016pixelwise}.
Learning-based approaches have also been incorporated for MVS to reconstruct 3D from few images~\cite{niemeyer2020differentiable,wiles2020synsin, yariv2020multiview, leroy2021volume, sitzmann2019srns,jang2021codenerf}, and more recently by adopting the RAFT architecture~\cite{cermvs} or using generic Transformers~\cite{cao2024mvsformer++}.
In contrast, the recent \duster{} framework~\cite{wang2024dust3r,leroy2024mast3r}, upon which we build our approach, departs from these methods altogether by defining a simple and unified neural framework to 3D reconstruction based on the paradigm of pointmap regression.
The idea is that, ideally, learning-based MVS should be capable of reconstructing 3D from a few images without strictly requiring the availability of camera parameters.

\myparagraph{Guiding 3D.} 
In this paper, we refer to \emph{guidance} as the notion that a method can  \emph{optionally} take in 3D priors (typically camera parameters or sparse depth) to help with some prediction task.
This can take different forms in the vast literature of 3D vision works.
COLMAP~\cite{schonberger2016pixelwise}, for instance, can take camera parameters for rectifying camera views; however, it is an optimization-based method, meaning that each scene has to be processed independently. %
Recently, UniDepth~\cite{piccinelli2024unidepth} is a monocular depth and intrinsic estimation method capable of optionally providing camera intrinsics as additional guide to generate depthmap, with an architecture specifically handcrafted for that purpose.
Also, depth completion models~\cite{guidenet,bpnet,yan2024tri,lrru} such as CompletionFormer~\cite{completionformer2023} take sparse depth input as guidance in order to inpaint the full depthmap, but are also typically capable of inferring depth from an image alone.
Overall, however, there is currently no learning-based model to the best of our knowledge that guides its output by taking any subset of sparse depth, camera intrinsics and camera extrinsics.

\myparagraph{Depth Completion.} 
Given sparse depths and a corresponding RGB image, the task is to densify  depth by propagating sparse measurements.
CNN-based approaches were adopted by ~\cite{ma2018self, ma2018sparse}, and ~\cite{cheng2018depth} proposed convolutional spatial propagation network (CSPN). 
Multi-branch networks, which process RGB and sparse depth separately, were used in ~\cite{hu2021penet, nazir2022semattnet, Qiu_2019_CVPR, guidenet, wvangansbeke2019sparse, zhang2018deepdepth, rho2022guideformer, yan2023rignet}.
CompletionFormer~\cite{completionformer2023} leverages both CNN and ViT architectures to capture both global and local features.
Most methods focus only on the specific depth completion task from a single image and partial depth as the input, often without incorporating camera intrinsics.
Using the pointmap representation, \ours{}, based on a ViT architecture, can complete the depth map from both RGB images and sparse point clouds, while taking relative pose or camera intrinsics as auxiliary input.

\myparagraph{Pose Estimation from Sparse views.} 
Learning-based approaches involve regressing relative pose, applying probabilistic methods, using energy-based or diffusion models~\cite{rockwell2022, zhang2022relpose, lin2024relpose++, balntas2017hpatches, wang2023posediffusion}.
RayDiffusion~\cite{raydiffusion} models camera rays instead of camera parameters, applying diffusion on rays, then recovers the camera pose using DLT~\cite{abdel1971direct}.
\duster{} discovers camera pose from pointmaps using PnP, while \ours{} predicts pointmaps in both camera coordinate frames, allowing the use of Procrustes alignment.

\myparagraph{\duster{} and its developments.} \duster{}~\cite{wang2024dust3r} does not require calibrated images for 3D reconstruction, can handle extreme viewpoint changes, and deals with both single and a few images within the same framework.
Its pointmap representation eases the training by not having to deal with camera parameters directly.
MASt3R~\cite{leroy2024mast3r} improves \duster{} by also predicting dense local features, trained with an additional matching loss.
Splatt3R~\cite{smart2024splatt3r} marries MASt3R with GS. %
Spann3R~\cite{wang2024spann3r} predicts pointmaps in global coordinate and  adopts an external spatial memory that tracks relevant 3D information.
MonST3R~\cite{zhang2024monst3r} further trains  \duster{} on dynamic scenes and generates time-varying dynamic point clouds, as well as camera intrinsics and poses from a dynamic video.
In this work, we take a different and complementary direction and instead improve the core capabilities of \duster{}, which could benefit all aforementioned works.

\begin{figure}
    \centering
    \includegraphics[width=\linewidth,trim=0 350 480 0,clip]{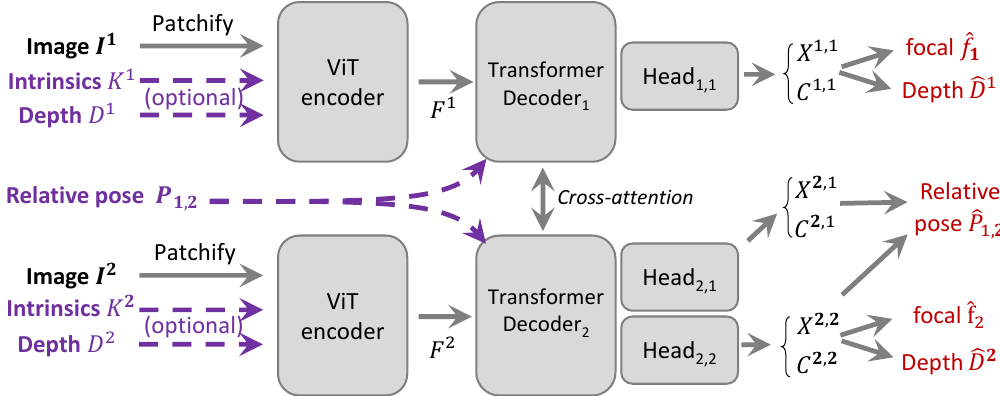} \\[-0.3cm]
    \caption{\textbf{Overview of the model architecture.} 
    Following \duster{}~\cite{wang2024dust3r}, images are encoded then decoded with a ViT backbone into pointmaps, from which focals, depthmaps and relative pose can be extracted.
    \ours{} introduces optional inputs to guide the regression with prior knowledge about the camera intrinsics and depth (fed into the encoder) and the pose (into the decoder).
    } 
    \label{fig:overview}
    \vspace{-0.2cm}
\end{figure}

\section{Methodology}
\label{sec:method}

We aim to train a network $\mathcal{F}$ that can take two input images $I^1,I^2\in \mathbb{R}^{W\times H\times 3}$ of a given static scene and any subset of auxiliary information $\Omega \subseteq \{K_1,K_2,P_{1,2},\D{1},\D{2}\}$, in order to regress a 3D reconstruction of the scene.
Here, $K_1,K_2\in\mathbb{R}^{3\times 3}$ are camera intrinsics, 
$P_{1,2}\in\mathbb{R}^{4\times 4}$ denote the relative pose between the two cameras, 
and $\D{1},\D{2} \in\mathbb{R}^{W\times H}$ are depthmaps with associated masks $M^1,M^2\in\{0,1\}^{W\times H}$ specifying pixels with valid depth data
(\ie masks are possibly sparse).
The network $\mathcal{F}$ is tasked to regress several pointmaps from which the camera intrinsic and extrinsic parameters as well as the dense depthmaps can be straightforwardly extracted for both images as described below. %

\myparagraph{Pointmaps.} 
For each pixel $(i,j)$ in an image $I$, we assume there exists a corresponding single 3D point $X_{i,j}$, where $X\in\R^{W\times H \times 3}$ is a \emph{pointmap}. 
Given camera intrinsics $K$ and a depthmap $D$, we can compute $X_{i,j}=K^{-1} [iD_{i,j}, jD_{i,j}, D_{i,j}]$ in the camera coordinate system.
In the following, we denote as $\X{n}{m}$ the pointmap of image $\I{n}$ expressed in the coordinate system of camera $\I{m}$. 
To swap the coordinate system from camera $\I{k}$ to camera $\I{m}$, we have $X^{n,k} = P_{m,k} X^{n,m}$ where $P_{m,k}=P_k P_m^{-1}$.

\subsection{Overall architecture}

We follow the recently proposed \duster{} framework~\cite{wang2024dust3r}.
\duster{} is a breakthrough foundation model based on vision transformers (ViT) that can regress two 3D pointmaps $\X{1}{1}, \X{2}{1}$ given solely two unposed and uncalibrated input images.
Nevertheless, \duster{}'s inability to leverage potentially available information about cameras or depth %
limits its practical use.
To remedy these drawbacks, we enhance the original \duster{} network $\mathcal{F}$ in two ways.

First, we introduce specific modules to incorporate any subset of extra information such as  camera intrinsics, camera poses and depthmaps seamlessly into \duster{}.
Second, we predict an additional pointmap $\X{2}{2}$, which represents the pointmap of image $\I2$ in its own coordinate system.
Predicting three pointmaps offers further capabilities, namely the possibility of recovering all information about \emph{both} cameras in a single forward pass, as explained in \cref{sec:applications}.
We first describe the overall architecture, see \cref{fig:overview}, and then detail the dedicated modules to inject auxiliary information.

\myparagraph{Encoder.} 
We use a shared ViT encoder to encode both images independently. 
In addition to $I$, the encoder can receive auxiliary information about intrinsics $K$ and depth $D$ for each image.
For the two input images $I^1$, $I^2$ and their respective auxiliary information $\Omega_1 \in \sigma(\{K_1,D_1\}), \Omega_2 \in \sigma(\{K_2,D_2\})$, where $\sigma$ denotes the set of all subsets, the encoder processes the information in a Siamese manner:
\begin{equation}
  F^1 = \text{Encoder}(I^1,\Omega_1), ~~ F^2 = \text{Encoder}(I^2,\Omega_2).
\end{equation}

\myparagraph{Decoder.}
Likewise, the network has two decoders, each with its corresponding head, one predicting $X^{1,1}$ and the other one estimating $X^{2,1}$ and $X^{2,2}$. 
Both decoders communicate via cross-attention between their own tokens and the outputs of the previous block of the other decoder. 
Each decoder may receive the relative pose $P_{12}$ as additional input or not. 
Assume we provide the auxiliary information $\Omega_D \in \sigma(\{P_{12}\})$ at the $i$-th block of both decoders:
\begin{eqnarray}
  G_i^1 = \text{DecoderBlock}^1_{i}\left(G_{i-1}^1, G_{i-1}^2, \Omega_D\right), \nonumber \\
  G_i^2 = \text{DecoderBlock}^2_{i}\left(G_{i-1}^2, G_{i-1}^1, \Omega_D\right),
\end{eqnarray}
After $B$ decoder blocks in each branch,
the head regresses several pointmaps and their associated confidence map:
\begin{eqnarray}
  X^{1,1}, C^{1,1} = \text{Head}^1\left(G_B^1\right), \nonumber \\
  X^{2,1}, X^{2,2}, C^{2,1}, C^{2,2} = \text{Head}^2\left(G_B^2\right).
\end{eqnarray}
Unlike \duster{} which uses a DPT head~\cite{dpt}, here we employ a linear head for both $\text{Head}^1$ and $\text{Head}^2$ %
as it is more efficient and performs similarly.

\myparagraph{3D Regression Loss.} 
Following \duster{}, we supervise the model to minimize the distance between ground-truth and predicted pointmaps in a scale-invariant manner, allowing the model to train on multiple datasets with various scales.
The regression loss between predicted and ground-truth pointmaps (resp. $\X{n}{m}$ and $\Xgt{n}{m}$) at pixel $(i,j)$ is defined as $\lreg_{i,j}(n,m)=\Vert \X{n}{m}_{i,j}/\z_m  - \Xgt{n}{m}_{i,j}/\zgt_m \Vert$, where $\z_m,\zgt_m$ serve as scale normalizer.
That is, $\z_m$ is the average norm of all valid 3D points expressed in coordinate system of image $\I{m}$, \ie  $\z_1=\Z(\X{1}{1} \cup \X{2}{1})$,  $\z_2=\Z(\X{2}{2})$ and likewise for $\zgt_1,\zgt_2$, with $\Z(X)=\text{mean}(\{\Vert X_{i,j} \Vert\, | i,j \in \valid_X\})$ and $\valid_X$ the set of valid pixels. 

\myparagraph{Confidence-aware loss.} 
The model jointly learns to predict a confidence level $\C{n,m}_{i,j}$ of each pixel $(i,j)$.
The confidence-aware regression loss for a given pointmap $\X{n}{m}$ can be expressed as the 3D regression loss $\lreg$ weighted by the confidence map:
\begin{equation}
  \lconf(n,m) = \sum_{i,j\in\valid} \C{n,m}_{i,j} \lreg_{i,j}(n,m) - \alpha \log \C{n,m}_{i,j},
\label{eq:conf_loss}
\end{equation}
with $\alpha=0.2$. This loss penalizes the model less when the prediction is not accurate on harder areas, encouraging the model to extrapolate.
The final loss is expressed as
\begin{equation}
  \mathcal{L} = \lconf(1,1) + \lconf(2,1) + \beta \lconf(2,2),
\end{equation}
where $\beta$ is a hyper-parameter typically set to $\beta=1$. 

\begin{figure}
    \centering
    \includegraphics[width=0.9\linewidth,trim=0 100 330 0,clip]{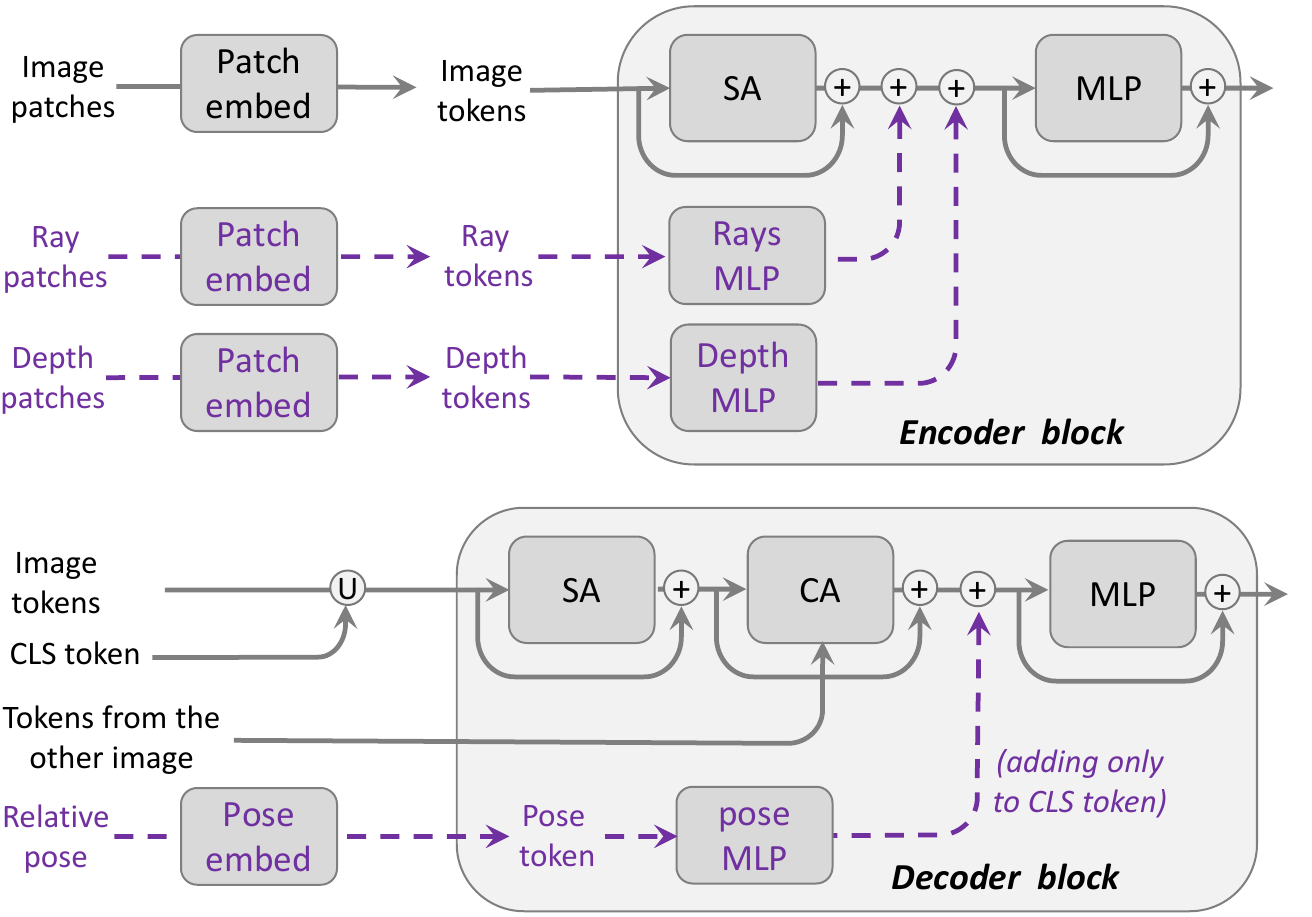} \\[-0.3cm]
    \caption{\textbf{Architecture for injecting auxiliary modalities.} \textit{Top:} injection of optional intrinsics and depth into the encoder. Intrinsics are encoded into ray patches, sparse depth is %
    patchified. Each of these modalities goes into a block-specific MLP and are token-wise added in the middle of the transformer block.
    \textit{Bottom:} injection of optional relative pose into the decoder. The relative pose is fed to a first embedding layer followed by an MLP. This token is added to the CLS token of the decoder after the self-attention (SA) and cross-attention (CA), but before the MLP.
    Experiments show that injection in the first block only suffices.}
    \vspace{-0.4cm}
    \label{fig:inject}
\end{figure}

\begin{figure*}
    \centering
    \includegraphics[width=1\linewidth, trim=0 320 0 0, clip]{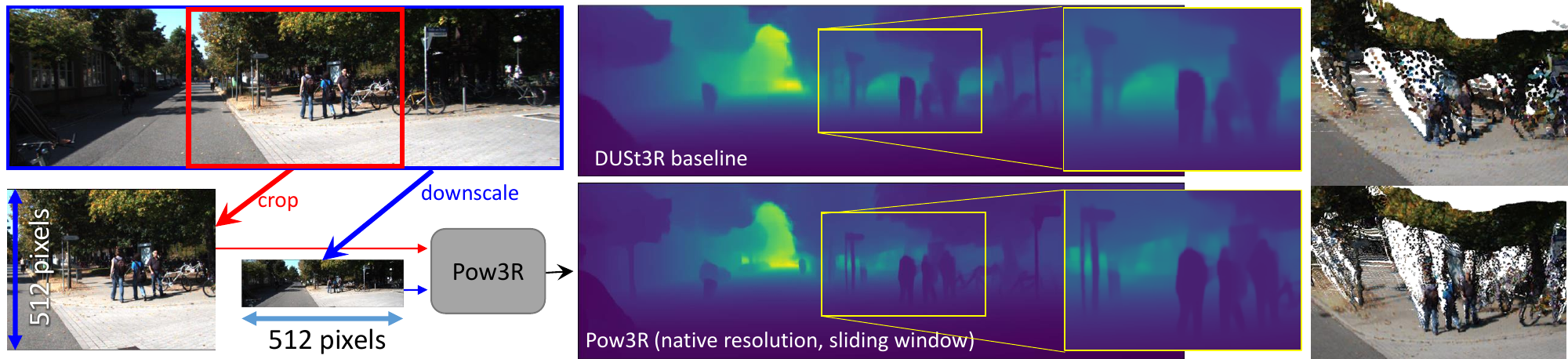} \\[-0.3cm]
    \caption{\textbf{Native resolution example.} Methods like DUSt3R require centered principal point and would thus downscale the image to its training resolution, leading to blurry prediction (top). With camera intrinsics input, we can process any crop in the image leading to native resolution prediction (bottom row), for example by dividing the image into 4 crops, that are independently process as the first input image, while the second input image of the network is set to the downscaled image.  
    }
    \vspace{-0.2cm}
    \label{fig:native_ex}  
\end{figure*}

\subsection{Adding Versatile Conditioning}

The knowledge of auxiliary information can significantly enhance 3D predictions at test time.
Our model leverages up to five different modalities, which include two intrinsics $K_1,K_2$, two depthmaps $D^1,D^2$ for each image and the relative pose $P_{12}$.
To condition the output on it, we first embed the auxiliary information using dedicated MLPs and then inject these embeddings at different points in the pipeline.

We discuss different options in this regard. 
In the first strategy, denoted as `\embed{}', we simply add the auxiliary embeddings to the token embeddings before the first transformer block, as would be done for standard positional embeddings~\cite{vit}.
In the second strategy, denoted as `\inject{n}', we instead insert dedicated MLPs for each modality in a subset of $n$ transformer blocks, as shown in \cref{fig:inject}.
Our experiments show that `\inject{1}' performs slightly better than `\embed{}' and similarly with `\inject{n}', where $n>1$, see \cref{sec:ablations}.
We now describe how to compute the initial embeddings for each specific modality.

\myparagraph{Intrinsics.} 
Following~\cite{raydiffusion}, we generate camera rays from the intrinsic matrix $K\in\R^{3\times 3}$, thereby establishing a direct correspondence between RGB pixels and rays. 
The ray at pixel location $(i,j)$ is computed as $K^{-1} [i, j, 1]$ and encodes the viewing direction of that pixel \wrt the current camera pose.
This allows us to potentially process non-centered crops and hence to perform inference in higher image resolutions, see \cref{sec:applications}.
Similarly to RGB inputs, we patchify and embed dense rays and provide them to the encoder.

\myparagraph{Depthmaps / Point Clouds.} 
Given a depthmap $\D{}$ and its sparsity mask $M$, we first normalize $D'=D/\Z(D)$ to handle any depth ranges at train and test time. 
As for RGB and rays, we patchify the stacked maps $[D',M]\in\R^{W\times H\times 2}$ and compute patch embeddings, which are then fed to the encoder.
By jointly patchifying the depth and its valid mask we can handle any level of sparsity. %

\myparagraph{Camera Pose.} 
Given the relative pose $P_{12}=[R_{12}|t_{12}]$, we normalize the translation scale as $t'_{12} = t_{12} / \Vert t_{12} \Vert$ since our output is unscaled. %
Unlike depthmaps or camera intrinsics, the camera pose cannot be expressed as a dense pixel map. %
Rather, camera pose affects the whole pixels between two images, so the embedding is instead added to the global CLS token of both decoders. %

\subsection{Downstream Tasks}
\label{sec:applications}

\myparagraph{Depthmaps.} 
In the pointmap representation, the $z$-axis of $\X{1}{1},\X{2}{2}$ directly corresponds to the depthmaps of the first and second image, respectively.
Note that \duster{} only outputs $\X{1}{1}$, hence it requires two forward passes of the pairs $(\I1, \I2)$ and $(\I2, \I1)$ to extract both depthmaps.

\myparagraph{High-resolution processing.}
A limitation of \duster{} is that it is trained for a specific resolution and does not generalize to larger resolutions.
A simple fix would be to include higher resolutions during training, but (i) many datasets do not provide high-resolution ground-truths, and (ii) the training cost could become prohibitive.
\ours{}, in contrast, can handle crops natively given camera intrinsics of the crop, as these provide the crop position information (\ie via focal length and principal point).
We can thus perform prediction in a sliding window fashion, yielding predictions matching any target resolution by simple stitching.
Comparison between standard and higher resolution processing are shown in \cref{fig:native_ex}.
Note that prediction for each crop may have a different scale, by design, and cannot be stitched directly.
We find that simply computing the median scale factor in overlapping areas is sufficient, and we blend the overlapping regions based on confidence without further post-processing.

\myparagraph{Focal Estimation.} 
Similar to \duster{}, we can recover focals for both input images from pointmap $\X{1}{1}$ and $\X{2}{2}$ with the robust Weiszfeld fast iterative solver~\cite{Plastria2011Weiszfeld}.
Similar to depthmap prediction, \duster{} needs to infer $(I_2,I_1)$ in order to compute the focal of the second image, while \ours{} predicts them in a single pass.

\myparagraph{Relative Pose Estimation.} \ours{} predicts the relative pose directly by  Procrustes alignment~\cite{Luo1999Procrustes, bregier2021deepregression} to get the scaled relative pose $P^{*}=[R^{*}|t^{*}]$ between $\X{2}{2}$ and $\X{2}{1}$, as it predicts the pointmaps of the second image in two different camera coordinates. 
{\small
\begin{equation}
   R^*, t^* {=} \argmin_{\sigma,R,t} \sum_{i,j} \sqrt{\C{2,2}_{i,j} \C{2,1}_{i,j}} \left\Vert \sigma (R \X{2}{2}_{i,j} {+} t) {-} \X{2}{1}_{i,j} \right\Vert^2. 
\end{equation}}
\noindent Procrustes alignment is known to be sensitive to noise and outliers; however, in our case, it performs similar to RANSAC~\cite{Fischler1981RANSAC} with PnP~\cite{hartley2003multiple,Lepetit2009EPnP}, while being an order of magnitude faster (see \cref{sec:pose}).

\myparagraph{Global alignment.}
The network $\mathcal{F}$ predicts pointmaps for image pairs.
To align all predictions in the same world coordinate system, we resort to the global alignment algorithm from \duster{}~\cite{wang2024dust3r} 
that minimizes a global energy function to find per-camera intrinsics, depthmaps and poses that are consistent with all the pairwise predictions.
Results of the optimization are global scene point-clouds, which can for instance serve for multi-view stereo estimation. %

\subsection{Training Procedure}

During training, we feed the network with annotated image  associated with random subsets of auxiliary information, the goal being for it to learn to handle any situation at test time.
For each pair, we first chose a random number $m$ of 
 modalities with uniform probability, and then randomly select the $m$ modalities likewise.
Depthmaps are randomly sparsified. %
When giving intrinsics, we also perform aggressive non-centered cropping with 50\% probability, so that the network learns to perform high-resolution inference. %

\myparagraph{Training data and recipe.} 
We follow the training recipe from \duster{}~\cite{wang2024dust3r} in terms of training data and protocol.
Namely, we use the provided pairs for 8 datasets: Habitat~\cite{savva2019habitat}, MegaDepth~\cite{li2018megadepth}, ARKitScenes~\cite{baruch2021arkitscenes}, Static Scenes 3D~\cite{mayer2016large}, BlendedMVS~\cite{yao2020blendedmvs}, ScanNet++~\cite{yeshwanth2023scannet++}, Co3D-v2~\cite{reizenstein2021common} and Waymo~\cite{sun2020scalability}, resulting in a total of 8.5M image pairs.
These datasets include indoor and outdoor scenes, as well as real and synthetic ones.
To train the model, we follow~\cite{wang2024dust3r} and first train the network with a resolution of 224px for 3 days on 8 A100 GPUs, and then finetune it in 512px with variable aspect-ratio for 2 more days.

\section{Experimental Evaluation}
\label{sec:xps}

\begin{table}[]
    \centering
    \definecolor{darkgreen}{HTML}{005701}
    \newcommand{\withmod}{\checkmark}
    \newcommand{\withoutmod}{$\times$}
    \newcommand{\firstcell}[1]{#1 & }
    \newcommand{\onediffcell}[2]{#1 & \textcolor{darkgreen}{\small (+#2)}}
    \resizebox{\linewidth}{!}{
    \begin{tabular}{c|c@{~~}c@{~~}c@{~~}c@{~~}c|r@{}rr@{}rr@{}rr@{}r}
\toprule
 & \multicolumn{5}{c|}{aux. modalities} & \multicolumn{2}{c}{focal} & \multicolumn{2}{c}{depth} & \multicolumn{4}{c}{rel. pose} \\
 & K1 & K2 & D1 & D2 & RT & \multicolumn{2}{c}{acc@1.015} & \multicolumn{2}{c}{$\tau$@1.03} & \multicolumn{2}{c}{RRA@2$^\circ$} & \multicolumn{2}{c}{RTA@2$^\circ$} \\
\midrule
DUSt3R & \withoutmod & \withoutmod & \withoutmod & \withoutmod & \withoutmod & \firstcell{36.6} & \firstcell{77.6} & \firstcell{69.2} & \firstcell{49.7} \\
\midrule
\multirow{12}{*}{\textbf{\ours{}}} & \withoutmod & \withoutmod & \withoutmod & \withoutmod & \withoutmod & \firstcell{39.4} & \firstcell{79.4} & \firstcell{71.9} & \firstcell{53.8} \\
 & \withmod & \withoutmod & \withoutmod & \withoutmod & \withoutmod & \onediffcell{75.4}{36.0} & \onediffcell{80.6}{1.2} & \onediffcell{77.1}{5.3} & \onediffcell{62.8}{8.9} \\
 & \withoutmod & \withmod & \withoutmod & \withoutmod & \withoutmod & \onediffcell{67.3}{27.9} & \onediffcell{80.3}{0.9} & \onediffcell{74.5}{2.6} & \onediffcell{59.1}{5.2} \\
 & \withmod & \withmod & \withoutmod & \withoutmod & \withoutmod & \onediffcell{98.0}{58.6} & \onediffcell{81.4}{2.0} & \onediffcell{80.3}{8.5} & \onediffcell{74.2}{20.3} \\
 & \withoutmod & \withoutmod & \withmod & \withoutmod & \withoutmod & \onediffcell{48.9}{9.5} & \onediffcell{89.1}{9.7} & \onediffcell{82.5}{10.6} & \onediffcell{64.9}{11.0} \\
 & \withoutmod & \withoutmod & \withoutmod & \withmod & \withoutmod & \onediffcell{49.5}{10.1} & \onediffcell{91.0}{11.7} & \onediffcell{83.4}{11.5} & \onediffcell{64.8}{10.9} \\
 & \withoutmod & \withoutmod & \withmod & \withmod & \withoutmod & \onediffcell{58.2}{18.9} & \onediffcell{\textbf{95.4}}{\textbf{16.0}} & \onediffcell{89.6}{17.7} & \onediffcell{77.6}{23.8} \\
 & \withoutmod & \withoutmod & \withoutmod & \withoutmod & \withmod & \onediffcell{48.6}{9.2} & \onediffcell{81.6}{2.2} & \onediffcell{92.3}{20.5} & \onediffcell{77.0}{23.1} \\
 & \withmod & \withmod & \withmod & \withmod & \withoutmod & \onediffcell{99.2}{59.8} & \onediffcell{\textbf{95.4}}{\textbf{16.0}} & \onediffcell{95.7}{23.9} & \onediffcell{94.3}{40.5} \\
 & \withmod & \withmod & \withoutmod & \withoutmod & \withmod & \onediffcell{98.1}{58.7} & \onediffcell{82.9}{3.6} & \onediffcell{95.1}{23.2} & \onediffcell{87.3}{33.5} \\
 & \withoutmod & \withoutmod & \withmod & \withmod & \withmod & \onediffcell{68.6}{29.2} & \onediffcell{\textbf{95.4}}{\textbf{16.0}} & \onediffcell{98.1}{26.2} & \onediffcell{91.3}{37.5} \\
 & \withmod & \withmod & \withmod & \withmod & \withmod & \onediffcell{\textbf{99.3}}{\textbf{59.9}} & \onediffcell{\textbf{95.4}}{\textbf{16.0}} & \onediffcell{\textbf{99.0}}{\textbf{27.2}} & \onediffcell{\textbf{98.1}}{\textbf{44.3}} \\
\bottomrule

    \end{tabular}
    }
    \vspace{-0.3cm}
    \caption{\textbf{Impact of guiding at test time} for models trained at $224{\times}224$ resolution. We report performances on Habitat for DUSt3R, which cannot handle auxiliary modalities, and our single model with different sets of modalities; we show in green the absolute improvement \wrt the results without auxiliary modality.
    }
    \vspace{-0.4cm}
    \label{tab:modality_study}
\end{table}

We first study the impact of auxiliary information guidance  (\cref{sec:xp_guiding}). 
We then experiment across many 3D vision tasks (\cref{sec:xp_mvd,sec:xp_mvds,sec:xp_mvstereo}) without retraining, \ie mostly zero-shot settings. We finally present some ablations in \cref{sec:ablations}.

\myparagraph{\duster{} baseline.}
We point out that our overall training data, protocols and architecture are exactly the same as \duster{}, except for the additional head that predict $\X{2}{2}$ (+0.1\% parameters) and for the modules used for injecting the auxiliary information (+4\% parameters for the \inject{1} variant that we use unless stated otherwise).

\myparagraph{Metrics.}
When evaluating focal lengths, we report the accuracy for a tight relative error threshold of 1.5\%.
For evaluating depthmap, we report the Absolute Relative Error (rel) and the Inlier Ratio ($\tau$) at a threshold of 3\%.
When evaluating poses, we opt for scale-invariant metrics and measure the relative rotation and translation accuracies (RRA and RTA resp.) of the relative pose at task-specific thresholds, and the mean Average Accuracy (mAA), defined as the area under the curve accuracy of the angular differences at min(RRA, RTA). %

\subsection{Guiding the output prediction}
\label{sec:xp_guiding}

\myparagraph{Impact of guidance.}
We first study the impact of providing different subsets of auxiliary information in terms of downstream 3D vision performance.
Namely, we report accuracies in \cref{tab:modality_study} for focal, depth, and rotation prediction on the Habitat validation dataset. %
In the supplementary material, we also provide results on the Infinigen dataset~\cite{Raistrick2023Infinigen}, which is not included in our training set, and observe similar outcomes.
We first note that in the absence of modalities, \ours{} performs slightly better than a \duster{} baseline, even though \ours{} is trained with just RGB images during a fraction of its training.
This shows that \ours{} does not lose its native strength for unconstrained regression despite having gained more capabilities.

Interestingly, we observe in \cref{tab:modality_study} the impact of each modality for each prediction task, and the synergetic effects of interacting modalities.
For instance, the model performs better for focal length estimation with depthmaps and/or relative pose, and \ours{}'s depth prediction improves with intrinsics and/or relative pose.
Moreover, relative pose prediction becomes more accurate with intrinsics and/or depthmap.
Overall, the model's performance consistently improves when provided with additional camera or scene priors, demonstrating the effectiveness of \ours in handling any combination of priors in a unified fashion.
We are not aware of any other approach that can seamlessly integrate all these priors in a learning-based method.

\begin{figure}
    \centering
    \includegraphics[width=\linewidth]{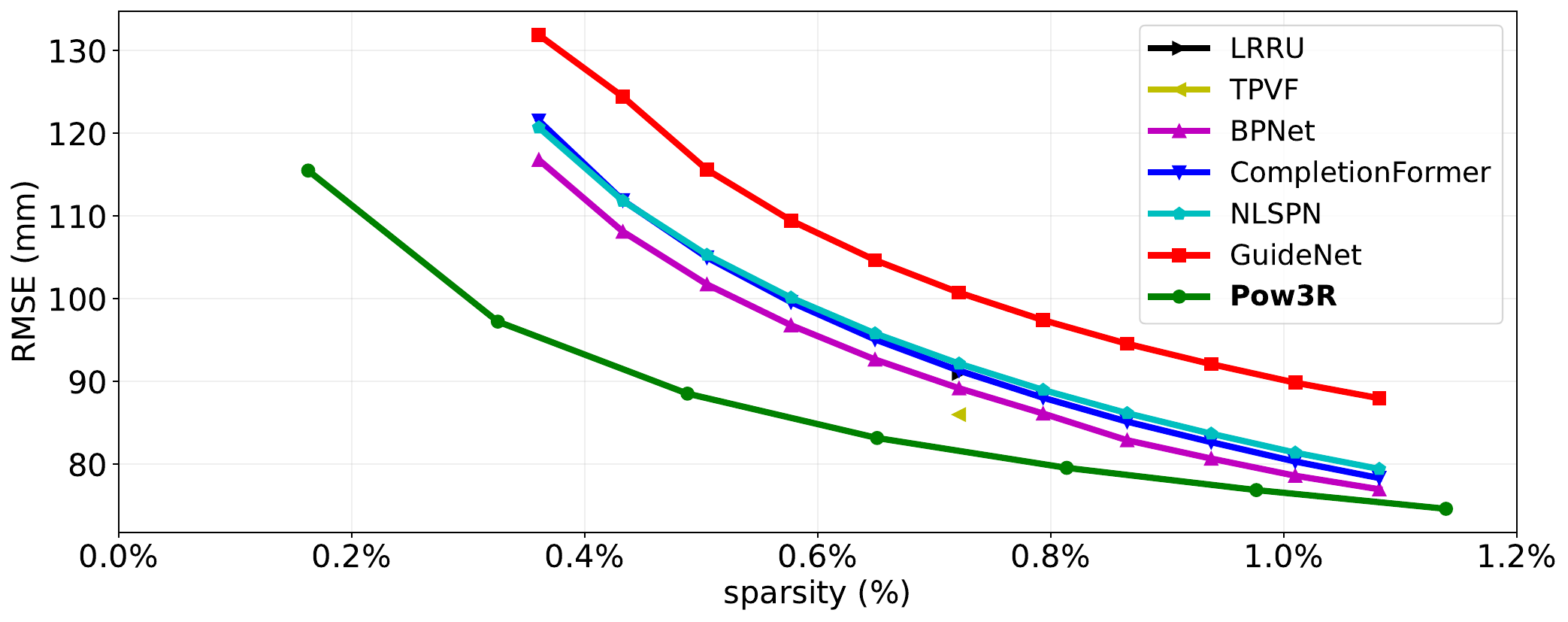} \\[-0.4cm]
    \caption{\textbf{Depth completion performance} on NYUd for various input depth sparsity. 
    }
    \vspace{-0.2cm}
    \label{fig:depthcompletion_nyu}
\end{figure}

\myparagraph{Depth completion.}
We also evaluate \ours{} on the depth completion task given RGB images with sparse depthmaps.
We vary the sparsity (the ratio of pixels with known depth) %
and demonstrate its impact on the NYUv2 dataset~\cite{SilbermanHKF12} in Figure~\ref{fig:depthcompletion_nyu}.
Although \ours{} is not trained on the NYUv2 dataset, it still achieves state-of-the-art results for sparse depth completion.
Our analysis indicates that the true performance of \ours{} on NYUv2  is likely underestimated due to numerous errors in the ground-truth depth annotations; further details are available in the supplementary material. %
On the KITTI~\cite{geiger2013vision} dataset, we obtain a mean absolute error below 30cm, which is competitive with the state of the art, 
considering that KITTI is not included in the training set.

\myparagraph{High-Resolution processing.}
With the possibility of specifying camera intrinsics, new capabilities arise.
Namely, \ours{} can perform high-resolution pointmap regression with a simple sliding-window approach.
We process fixed-resolution crops of the original images under the guidance of both (i) the crop information, provided in term of camera intrinsics, and (ii) a second image, itself possibly downscaled or cropped as well.
More details are provided in \cref{sec:applications} and \cref{fig:native_ex}.
When applied, results significantly improve compared to processing the images at 512-pixel resolution, as shown in \cref{tab:native}.
We also compare with a naive baseline where we simply feed high-resolution images directly to \ours{}, but it dramatically fails.

\begin{table}[]
    \centering
    \resizebox{\linewidth}{!}{
    \begin{tabular}{lc@{~}cc|cccccc}
    \toprule 
     & \multicolumn{2}{c}{\small{aux. mod.}} & High- & \multicolumn{2}{c}{KITTI} & & \multicolumn{2}{c}{T\&T} & Time \\
    \cmidrule{5-6} \cmidrule{8-9} 
     & Ks & RT & Res. & rel$\downarrow$ & $\tau\uparrow$ &  & rel$\downarrow$ & $\tau\uparrow$ & (sec)$\downarrow$\\
    \midrule
    \duster{}512 & $\times$ & $\times$ & $\times$ & 5.4 & 49.5 &  & 3.3 & 75.1 & 0.13\\
    \midrule
    \ours{}512 & $\checkmark$ & $\checkmark$ & $\times$ & 5.3 & 48.7 &  & 3.2 & 78.2 & 0.13\\
    \ours{}512 & $\checkmark$ & $\checkmark$ & (n) & 7.5  & 34.4 &  & 3.9 & 68.0 & 0.48\\
    \ours{}512 & $\checkmark$ & $\checkmark$ & $\checkmark$ & \textbf{4.6} & \textbf{53.5} &  & \textbf{2.5} & \textbf{82.3} & 0.40\\
    \bottomrule
    \end{tabular}} \vspace{-0.3cm}
    \caption{\textbf{Multi-view depth estimation results in high-resolution}
        using a sliding-window scheme and providing the crop intrinsics to \ours{}. 
        (n): naively inputting high-resolution images without downsampling to \ours{}.
        }
    \vspace{-0.4cm}
    \label{tab:native}
\end{table}

\begin{figure}
    \centering
    \includegraphics[width=0.49\linewidth]{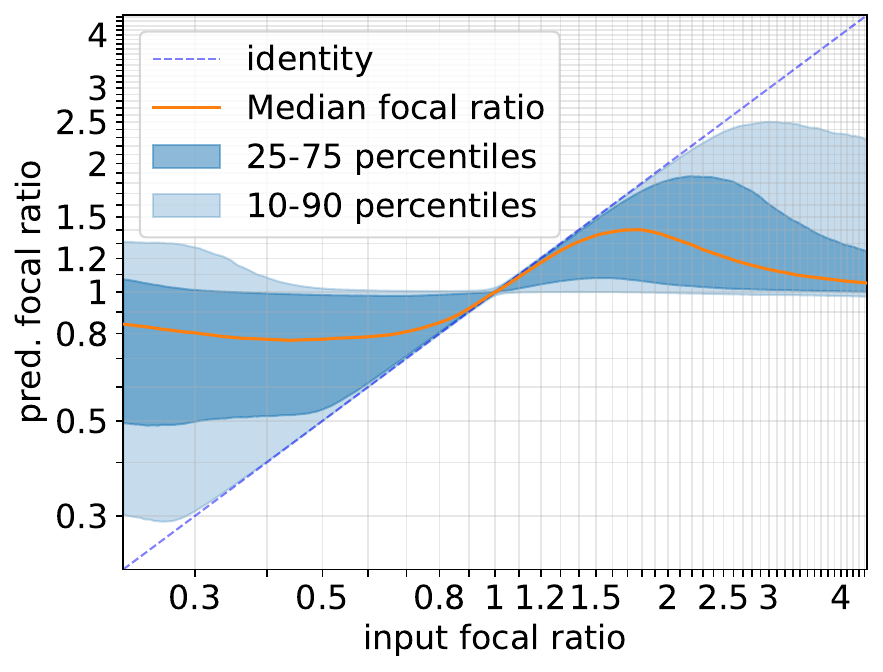} 
    \includegraphics[width=0.49\linewidth]{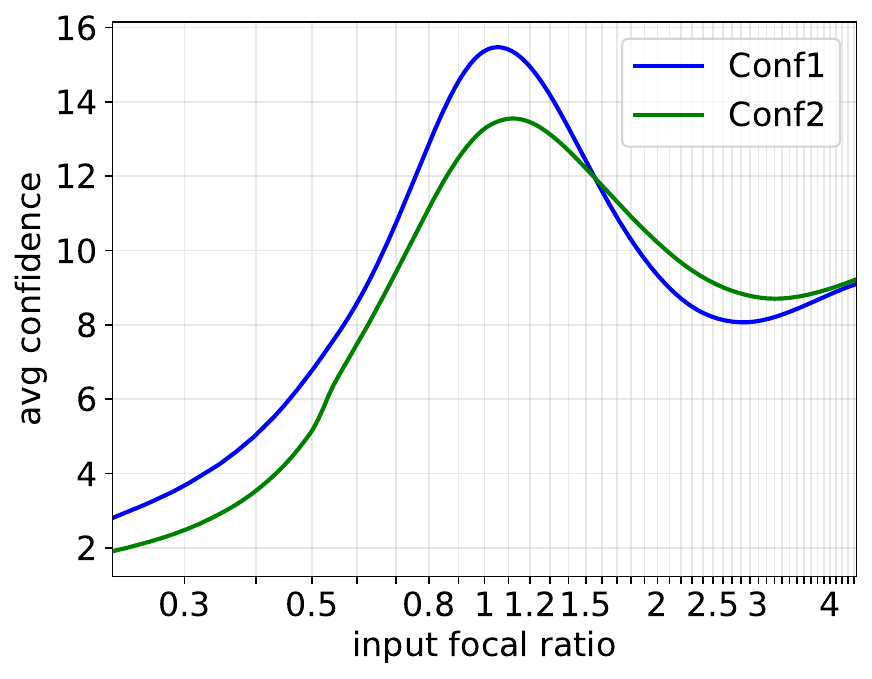} 
    \includegraphics[width=0.49\linewidth]{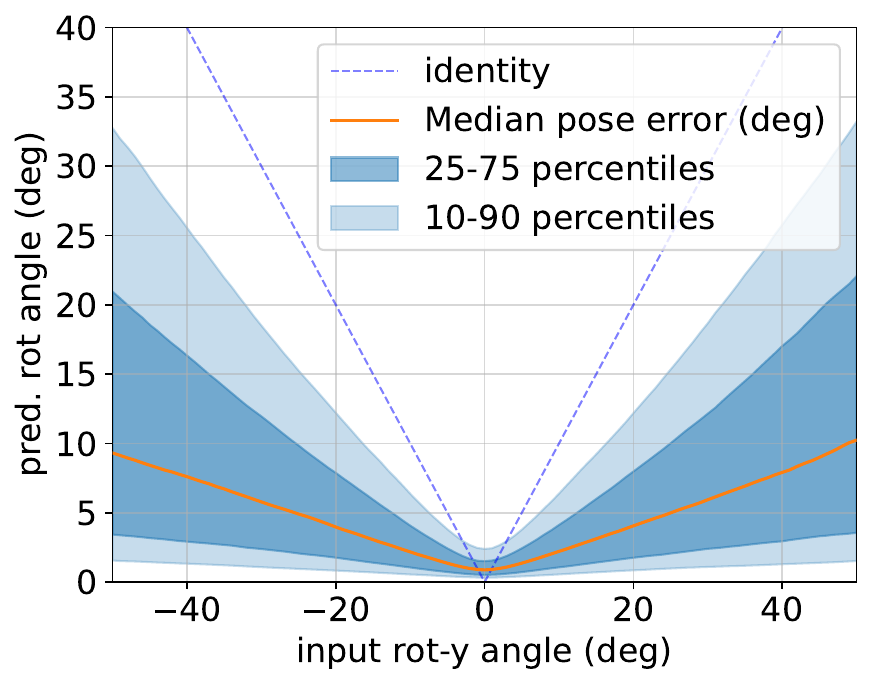}
    \includegraphics[width=0.49\linewidth]{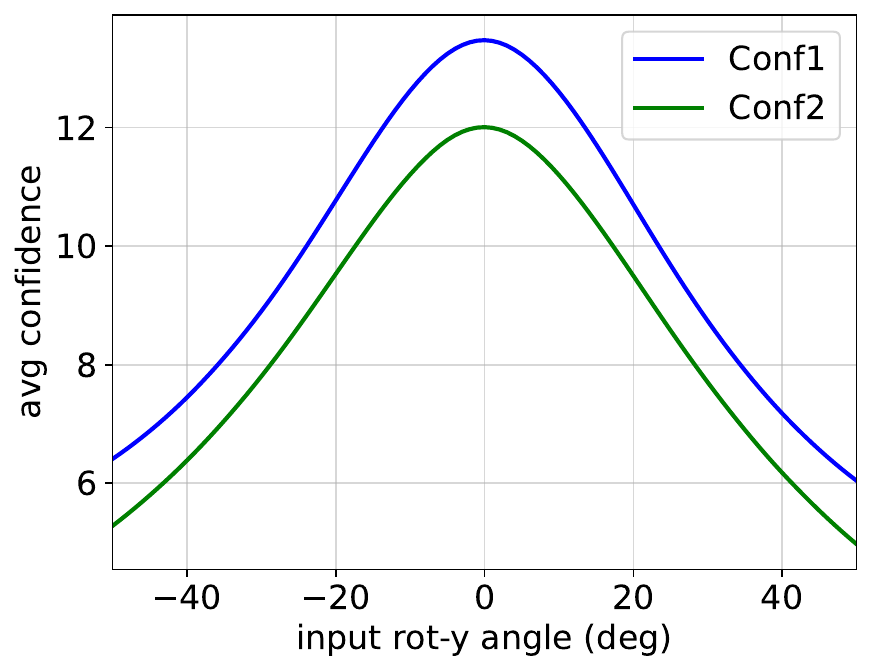}
    \\[-0.35cm]
    \caption{\textbf{Controllability study}.
        We study how the model reacts when presented with false focal (top) and false relative pose (bottom). We plot the input \emph{vs.} predicted camera parameters in the left column and the average confidence in the right column. Interestingly, the model seems to not take the input information at face value and instead makes an informed decision whether to use it or not, in which latter case the confidence becomes low.
        } 
    \label{fig:controlability}
    \vspace{-0.25cm}
\end{figure}

\myparagraph{Camera controllability.}
We study how \emph{controllable} are the regression outputs on 10K Habitat test pairs given arbitrary camera parameters.
For each test pair with ground-truth focal $f$ and relative pose $P_{12}$, we purposefully input a \emph{wrong} focal (resp. relative pose) and measure the output focal $\hat{f}$ (resp. pose $\hat{P}_{12}$) as predicted from the raw pointmaps.
Results are presented in \cref{fig:controlability} in terms of focal ratio $\nicefrac{\hat{f}}{f}$ and angle $d_{\text{geodesic}}(P_{12}, \hat{P}_{12})$.
Interestingly, the model accepts to follow the guidance whenever it is close to the ground truth value (\ie staying close to the identity transformation) %
but then refuses to follow it when it gets too different, showing that the model is able to somehow `judge' of the guidance's worth.
In all cases, the output confidence clearly correlates with the agreement between the guidance and the ground truth.
Overall, the model is thus not taking the auxiliary information at face value, and instead combines it with the other modalities (\eg RGB) it in a more subtle and informed way than what could be achieved with handcrafted methods.
This suggests that \ours{} would be robust to noise in the auxiliary information.

\subsection{Multi-View Depth Estimation}
\label{sec:xp_mvd}

Following RobustMVD~\cite{schroppel2022benchmark}, performances are measured on the KITTI~\cite{geiger2013vision}, ScanNet~\cite{dai2017scannet}, ETH3D~\cite{schops2017multi}, DTU~\cite{aanaes2016large}, and Tanks and Temples~\cite{knapitsch2017tanks} datasets. %
We report in \cref{tab:mvd} the Absolute Relative Error (rel) and Inlier Ratio ($\tau$) for each dataset when using (1) images alone, (2) images with relative poses, (3) images with camera intrinsics, and (4) images with both 
 modalities, and provide results for more state-of-the-art methods in the supplementary material. %

\myparagraph{Results}.
\ours{} again performs roughly on par with \duster{} in the absence of auxiliary information (\eg 3.73 average error across all datasets for \duster{}-512 \vs 3.64 for \ours{}-512). 
However, it is clear that providing auxiliary information consistently improves the depth estimates. %
In fact, \ours{}-512 with both pose and intrinsics achieves state-of-the-art performance on the ETH3D, DTU, and Tanks and Temples datasets, as well as on average across datasets.
Interestingly, intrinsics tend to improve performance more than the GT pose, which alone seems to have a negligible impact. 
This might be due to the fact that camera pose primarily helps the prediction of $\X{2}{1}$, while the depthmap of this task is obtained from the $z$ axis of $\X{1}{1}$ alone.
Still, we observe a clear synergy in providing jointly intrinsics \emph{and} relative pose.

\begin{figure*}
\begin{minipage}{0.65\linewidth}
\resizebox{\linewidth}{!}{
\begin{tabular}{l@{}c@{~~~}|r@{~~~}rr@{~~~}rr@{~~~}rr@{}rr@{~~~}r|r@{~~~}r}
\toprule
\multirow{2}{*}{Method} & GT & \multicolumn{2}{c}{KITTI} & \multicolumn{2}{c}{ScanNet} & \multicolumn{2}{c}{ETH3D} & \multicolumn{2}{c}{DTU} & \multicolumn{2}{c|}{T\&T} & \multicolumn{2}{c}{\bf{Average}} \\
 & range & rel$\downarrow$ & $\tau\uparrow$ & rel$\downarrow$ & $\tau\uparrow$ & rel$\downarrow$ & $\tau\uparrow$ & rel$\downarrow$ & $\tau\uparrow$ & rel$\downarrow$ & $\tau\uparrow$ & rel$\downarrow$ & $\tau\uparrow$ \\
\midrule
COLMAP~\cite{schonberger2016structure,schonberger2016pixelwise} (K+RT) & $\times$ & 12.0 & {\bf 58.2} & 14.6 & 34.2 &{ 16.4}&{55.1}&{\bf 0.7}&{\bf 96.5}&{\bf 2.7}& {\bf 95.0} &{ 9.3} & { 67.8} \\
{\small COLMAP Dense~\cite{schonberger2016structure,schonberger2016pixelwise} (K+RT)} & $\times$ & 26.9 & 52.7 & 38.0 & 22.5 & 89.8 & 23.2 & 20.8 & 69.3 & 25.7 & 76.4 & 40.2 & 48.8 \\

MVSNet~\cite{yao2018mvsnet} (K+RT) & $\checkmark$ & 18.6 & 30.7 & 22.7 & 20.9 & 21.6 & 35.6 & (1.8) & $(86.7)$ & 6.5 & 74.6 & 14.2 & 49.7 \\

Vis-MVSNet~\cite{zhang2020visibility} (K+RT) & $\checkmark$ & 9.5 & \un{55.4}& 8.9 & 33.5 &{ 10.8}& 43.3 &{(1.8)} &{(87.4)} &{4.1}& \un{87.2} &{7.0} &{61.4} \\

MVS-Former++~\cite{cao2024mvsformer++} (K+RT) & $\checkmark$ & 29.2 & 15.2 & 15.2 & 21.9 & 21.4 & 32.5 & ({1.2}) & ({ 91.9}) & 7.6 & 71.5 & 14.9 & 46.6  \\
CER-MVS~\cite{cermvs} (K+RT) & $\times$ & 14.3 & 32.2 & 21.1 & 24.3 & 11.7 & 47.5 &  4.1 & 71.3 &  6.4 &  82.1 & 11.5 & 51.5 \\

\midrule
 \duster{}~\cite{wang2024dust3r} & $\times$ &               5.4 & 49.5 & \bf (3.1) & \bf (71.8) & 3.0 & \un{76.0} & 3.9 & 68.6 & 3.3 & 75.1 & 3.7 & 68.2 \\

 \textbf{\ours{}} & $\times$ &                5.7 & 45.7 & (3.2) & (68.8) & 3.0 & 74.7 & 3.0 & 74.3 & 3.3 & 76.6 & 3.6 & 68.0 \\
 \textbf{\ours{} w/ RT} & $\times$ &       5.7 & 45.8 & (3.2) & (69.7) & \un{2.9} & 75.6 & 3.3 & 71.6 & \un{3.2} & 77.9 & 3.7 & 68.1 \\
 \textbf{\ours{} w/ K} & $\times$ &  \bf 5.3 & 48.3 & (\textbf{3.1}) & (70.8) & \un{2.9} & \un{76.0} & 1.6 & 89.9 & \un{3.2} & 77.3 & \bf 3.2 & \un{72.5} \\
 \textbf{\ours{} w/ K+RT} & $\times$ &   \textbf{5.3} & {48.7} & (\textbf{3.1}) & (\un{71.4}) & \textbf{2.8} & \textbf{77.1} & \un{1.5} & \un{91.1} & \un{3.2} & 78.2 & \bf{3.2} & \textbf{73.3} \\

\bottomrule
\end{tabular}}
\vspace{-0.3cm}
\normalsize
\captionof{table}{
\textbf{Multi-view depth evaluation:} \ours{} with both pose and intrinsics performs better than \duster{}. More comparisons with the state of the art can be found in the supplementary material.
(Parentheses) denote training on data from the same domain. 
}
\label{tab:mvd}

\end{minipage} \hfill
\begin{minipage}{0.33\linewidth}
\resizebox{\linewidth}{!}{
\begin{tabular}{lccc}
\toprule
Method & Acc.$\downarrow$ & Comp.$\downarrow$ & Overall$\downarrow$       \\
\midrule
\duster{}$^\dagger$~\cite{wang2024dust3r} &   2.677  &  \textbf{0.805}  & 1.741  \\ %
\duster{} (repr.) &   2.191  &  1.598  & 1.894  \\ %
{\bf \ours } &  2.116  &  1.370  & 1.743  \\ %
{\bf \ours w/ K} &  1.722  &  1.119  & 1.420  \\ %
{\bf \ours w/ Rt } &  2.205 & 1.429 & 1.817  \\ %
{\bf \ours w/ K+RT } &  \textbf{1.384} & 0.846 & \textbf{1.115}  \\ %
\bottomrule
\end{tabular}
}
\normalsize
\vspace{-0.3cm}
\captionof{table}{
\textbf{MVS results} with the accuracy, completeness and their average (in mm) on the DTU dataset. 
\duster{}$^\dagger$ means results from the original paper, and `repr.' means reproduced with the official public codebase and checkpoint.
\label{tab:mvs_dtu}
}

\end{minipage}
\vspace{-0.3cm}
\end{figure*}

\subsection{Multi-View Stereo}
\label{sec:xp_mvstereo}

We evaluate the quality of global 3D reconstructions on the DTU~\cite{aanaes2016large} benchmark in the presence and absence of auxiliary information.
For each subset of available information from $\{(K_1,K_2),P_{12}\}$, we extract pairwise predictions from image pairs and align them in a global coordinate system, as described in Sec \ref{sec:applications}.
Performance is evaluated in terms of accuracy, which is the smallest Euclidean distance to the ground-truth, and completeness as the smallest distance to the reconstructed shape, with the overall average.

\myparagraph{Results} 
are presented in \cref{tab:mvs_dtu}. 
We first attempt at reproducing the original \duster{} results using the official codebase and obtain slightly worse results than reported in~\cite{wang2024dust3r}.
Using the same codebase, \ours{} with RGB only is able to improve over these and match the original results.
Interestingly, adding ground-truth intrinsics results in a significant boost (19\% rel. improvement), but adding the camera poses instead slightly \emph{degrades} the results (-4\%).
However, using both intrinsics and pose results in a major improvement (36\% rel. improvement), demonstrating a synergetic effect between both modalities.
We hypothesize that the pose is of little value for easy cases like DTU, and that it does not help the model to properly resolve the focals.

\begin{table}[]
    \centering
    \resizebox{\linewidth}{!}{\begin{tabular}{c@{~~}l@{~}c@{~}cc@{~~}c@{~~}cccc}
\toprule 
\multirow{2}{*}{} & \multirow{2}{*}{Method} & GT &  & \multicolumn{3}{c}{Co3Dv2} &  & \footnotesize RealEstate10K & Speed\\
\cmidrule{5-7} \cmidrule{9-9} 
 &  & intrinsics &  & {\small RRA@15} & {\small RTA@15} & {\small mAA(30)} &  & mAA(30) & (fps)\\
\midrule 
\multirow{7}{*}{(a)} & Colmap+SG~\cite{sarlin2019coarse} & \checkmark &  & 36.1 & 27.3 & 25.3 &  & 45.2 & -\\
 & PixSfM~\cite{lindenberger2021pixel} & \checkmark &  & 33.7 & 32.9 & 30.1 &  & 49.4 & -\\
 & RelPose~\cite{zhang2022relpose} & $\times$ &  & 57.1 & - & - &  & - & -\\
 & PoseDiff~\cite{wang2023posediffusion} & $\times$ &  & 80.5 & 79.8 & 66.5 &  & 48.0 & -\\
 & RelPose++~\cite{lin2024relpose++} & $\times$ &  & 85.5 & - & - &  & - & -\\
 & RayDiff~\cite{raydiffusion} &  $\times$ &  & 93.3 & - & - &  & - & -\\
\midrule 
\multirow{4}{*}{(b)} & \duster{} (PnP)~\cite{wang2024dust3r} & $\times$ &  & 94.3 & 88.4 & 77.2 &  & 61.7 & 3.2\\
\cmidrule(lr){2-10}
 & \ours{} (PnP) & $\times$ &  & 94.8 & 89.9 & 78.5 &  & 62.5 & 3.2\\
 & \ours{} (Pro) & $\times$ &  & 94.6 & 90.3 & 78.1 &  & 66.3 & \textbf{30.9}\\
 & \ours{} w/ K (Pro) & \checkmark &  & \textbf{95.0} & \textbf{92.1} & \textbf{82.2} &  & \textbf{72.5} & 30.1\\
\bottomrule 
\end{tabular}

}
    \vspace{-0.3cm}
    \caption{\textbf{Multiview pose estimation on the CO3Dv2 and RealEstate10K.} 
    (a) denotes multiview methods, while (b) denotes pairwise methods.
    \ours{} with Procrustes alignment (Pro) performs better than \duster{} (PnP) and other multi-view based pose estimation approaches, and even more so with known intrinsics, while being much faster. 
    RealEstate10K is not in the training set.
    }
    \label{tab:multiview}
    \vspace{-0.5cm}
\end{table}

\subsection{Multi-View Pose estimation}
\label{sec:pose}
\label{sec:xp_mvds}

We then evaluate \ours{} for the task of multi-view pose prediction on the Co3Dv2~\cite{reizenstein2021common} and RealEstate10K~\cite{realestate10K} datasets. 
Co3Dv2 comprises over 37,000 videos across MS-COCO 51 categories with ground-truth camera derived from running COLMAP on 200 frames per video.
RealEstate10k contains 80,000 YouTube video clips, containing indoor/outdoor scenes, with camera poses obtained via SLAM followed by Bundle Adjustment.
For Co3Dv2, we follow the protocol of ~\cite{wang2023posediffusion} and for RealEstate10K, we use its test set of 1,800 videos. 
For each sequence, we randomly select 10 frames and feed 45 pairs to \ours{}.
We report the results in terms of RRA, RTA and mAA in \cref{tab:multiview}.

\myparagraph{Results.} \ours{} predicts $\X{2}{2}$, allowing to estimate the relative pose between 2 images in a single forward using Procrustes Alignment (see \cref{sec:applications}). %
In contrast, \duster{} does not output $\X{2}{2}$ and relies on PnP to obtain camera poses.
\ours{} with PnP achieves slightly better results than \duster{} with PnP.
Using Procrustes alignment yields similar results, while being orders of magnitude faster than RANSAC, with a roughly $10\times$ speed-up overall when combining inference and pose estimation times.
Finally, performance improves even further with known intrinsics, significantly outperforming all the state of the art.

\subsection{Architecture ablation}
\label{sec:ablations}

We evaluate the impact of adding the $\X{2}{2}$ prediction in \cref{tab:ablation}.
For these ablations, models are trained on a subset of Habitat and tested on focal, depth, and relative pose prediction, with results averaged over all combinations of auxiliary modalities.
\duster{} serves as a baseline, and \duster{}-ft is finetuned with an additional $\X{2}{2}$ head.
This output can be used at test time to directly predict focals or relative poses, within a single forward pass, instead of having to forward both the original pair and its swapped version.
We also report the performance, with and without predicting $\X{2}{2}$ and/or using it.
In all cases, predicting $\X{2}{2}$ leads to slightly better performance while evaluations can be performed in a single forward pass.

\begin{table}[]
    \centering
    \resizebox{\linewidth}{!}{
    \begin{tabular}{l@{~}cc|ccc@{~~}cc@{~~}c|c}
    \toprule
   \multirow{2}{*}{Method} & with & {\small Single} & Focal & Depth & \multicolumn{2}{c}{RRA@2} & \multicolumn{2}{c|}{RTA@2} & \multirow{2}{*}{Avg.} \\
    & $X^{2,2}$ & {\small Forward} & {\small acc@1.015} & $\tau$@1.03 & (Pro) & (PnP) & (Pro) & (PnP) & \\
    \midrule
    DUSt3R & $\times$ & $\times$ & 36.8 & 85.4 & 69.2 & 72.4 & 49.8 & 57.3 & 60.5 \\
    \multirow{2}{*}{DUSt3R-ft} & \checkmark & $\times$ & 38.9 & 79.1 & 71.7 & 74.4 & 52.9 & 59.2 & 62.7 \\
    & \checkmark & \checkmark & 38.3 & 78.5 & 73.1 & 74.4 & 56.3 & 59.2 & 63.3 \\
    \midrule
    \multirow{3}{*}{\textbf{\ours{}}} & $\times$ & $\times$ & 66.0 & 85.7 & 81.9 & 83.8 & 68.6 & 73.3 & 76.5 \\
     & \checkmark & $\times$ & \textbf{67.2} & \textbf{86.1} & 82.9 & 84.7 & 70.0 & \textbf{74.6} & 77.6 \\
     & \checkmark & \checkmark & 64.8 & 85.0 & \textbf{84.2} & \textbf{84.7} & \textbf{73.0} & \textbf{74.6} & \textbf{77.7} \\
    \bottomrule
    \end{tabular}}
    \vspace{-0.3cm}    
    \caption{\textbf{Impact of regressing $X^{2,2}$.} %
    Metrics (in \%)
     on Habitat
    are averaged over all combinations of input modalities.
    The `with $X^{2,2}$' column denotes methods trained to predict it. `Single forward' denotes evaluations of models based on $X^{2,2}$ prediction, \ie, in a single forward pass, instead of forwarding image pairs and their swapped versions. %
    The top part shows the original DUSt3R, or finetune (-ft) with $X^{2,2}$ prediction.
    }
    \label{tab:ablation}
    \vspace{-0.1cm}
\end{table}

\begin{table}[]
    \centering
    \resizebox{\linewidth}{!}{
    \begin{tabular}{lc|ccc@{~~}cc@{~~}c|c}
    \toprule
   \multirow{2}{*}{Arch.} & Extra & Focal & Depth & \multicolumn{2}{c}{RRA@2} & \multicolumn{2}{c|}{RTA@2} & \multirow{2}{*}{Avg.} \\
    & Params & {\small acc@1.015} & $\tau$@1.03 & (Pro) & (PnP) & (Pro) & (PnP) & \\
    \midrule
    Embed & 16.0M & 64.7 & 84.7 & 84.1 & 84.5 & 72.8 & 73.8 & 77.4 \\
    \textbf{Inject1} & 22.6M & 64.8 & 85.0 & 84.2 & 84.7 & 73.0 & 74.6 & 77.7 \\
     Inject4 & 42.3M & \textbf{65.9} & \textbf{85.1} & \textbf{84.8} & \textbf{85.2} & \textbf{73.7} & \textbf{75.4} & \textbf{78.3} \\  
    \bottomrule
    \end{tabular}}
    \vspace{-0.3cm}    
    \caption{\textbf{Architecture ablation}. %
    Results (in \%), on Habitat, 
    are averaged over all combinations of input modalities.
    We compare the injection architecture for different number of blocks with the `embed' architecture in which the extra signals are added before the transformer blocks. `Inject1' offers a good compromise between performance and number of extra parameters. 
    }
    \label{tab:ablation_arch}
    \vspace{-0.3cm}
\end{table}

We also explore different ways of conditioning auxiliary information in \cref{tab:ablation_arch}.
To incorporate additional priors, \ours{} injects those embeddings into the first block of the encoder or decoder, see \cref{fig:inject}.
We compare this approach to another design choice, where the model embeds them before going through attention blocks.
We have also tried to perform injection in more than one block, without significant improvement while the computational cost increases.

\section{Conclusion}

We have presented \ours{}, a novel approach to guide large 3D vision models, \eg~\duster{}, with camera and scene priors in an implicit manner.  
Not only does it results in consistent gains whenever auxiliary information is available, but it also unlocks new capabilities such as high-resolution processing and sparse-to-dense depth completion.

{
    \small
    \bibliographystyle{ieeenat_fullname}
    \bibliography{main}
}

\appendix
\renewcommand{\thefigure}{\Alph{figure}}
\renewcommand{\thetable}{\Alph{table}}
\setcounter{figure}{0} 
\setcounter{table}{0}  

\twocolumn[{%
\renewcommand\twocolumn[1][]{#1}%
{\Huge \textbf{Appendix}}
\vspace{-0.6cm}
\begin{center}
	\vspace{0.5cm}
    \captionsetup{type=figure}
    \includegraphics[width=\linewidth,trim=0 115 195 0,clip]{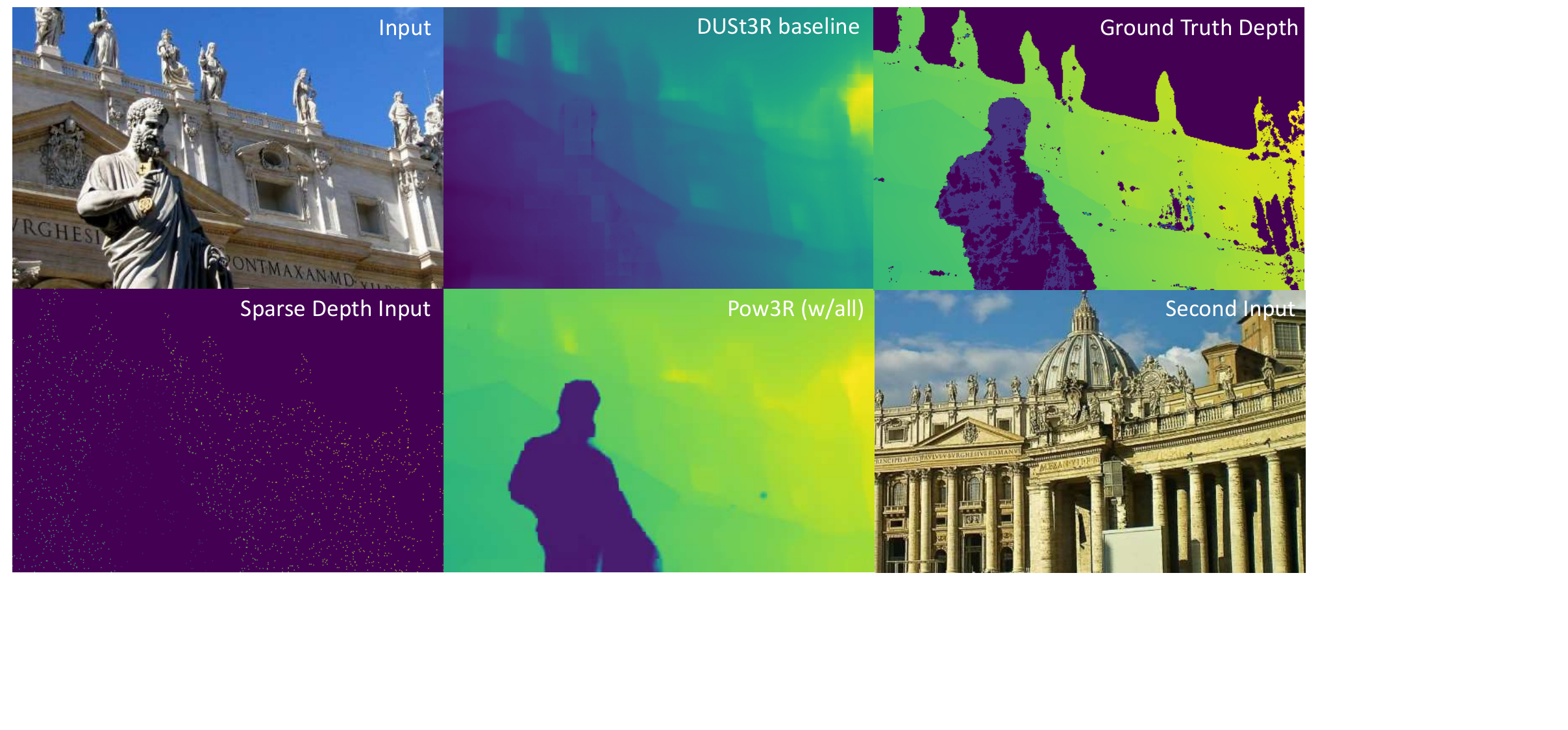}
    \vspace{-0.65cm}
    \captionof{figure}{\textbf{Qualitative Result in terms of predicted depthmaps.} We compare \ours{} with \duster{} on one of the Megadepth~\cite{li2018megadepth} outdoor scenes (from the validation set). For this evaluation, we feed camera intrinsics, pose as well as 2048 sparse point clouds. The figure clearly demonstrates that \duster{} fails to accurately  capture the statue, while \ours{} reconstructs it correctly.}
    \label{fig:qualitative_megadepth_statue}
\end{center}
}]

\section{Further Experiments on the Impact of Guiding}

In Table 1 of the main paper, we perform experiments on the Habitat validation dataset to study the impact of each auxiliary modality on the reconstruction accuracy in terms of different metrics. 
In Table~\ref{tab:modality_study_infinigen}, we perform the same experiment on data generated with Infinigen~\cite{Raistrick2023Infinigen}. 
Compared to Habitat, InfiniGen offers two advantages: 
(i) InfiniGen is free of artifacts, which can be numerous in the Habitat dataset due to acquisition problem on windows, mirrors, complex surfaces, \etc;
(ii) there is no risk of training data contamination or overfitting, as InfiniGen is not part of the training set at all.
Nevertheless, we observe similar results and trends than on Habitat, highlighting the robustness of these findings.

\begin{table}[]
    \centering
    \definecolor{darkgreen}{HTML}{005701}
    \newcommand{\withmod}{\checkmark}
    \newcommand{\withoutmod}{$\times$}
    \newcommand{\firstcell}[1]{#1 & }
    \newcommand{\onediffcell}[2]{#1 & \textcolor{darkgreen}{\small (+#2)}}
    \resizebox{\linewidth}{!}{
    \begin{tabular}{c|c@{~~}c@{~~}c@{~~}c@{~~}c|r@{}rr@{}rr@{}rr@{}r}
    \toprule
     & \multicolumn{5}{c|}{aux. modalities} & \multicolumn{2}{c}{focal} & \multicolumn{2}{c}{depth} & \multicolumn{4}{c}{rel. pose} \\
     & K1 & K2 & D1 & D2 & RT & \multicolumn{2}{c}{acc@1.015} & \multicolumn{2}{c}{$\tau$@1.03} & \multicolumn{2}{c}{RRA@2$^\circ$} & \multicolumn{2}{c}{RTA@2$^\circ$} \\
    \midrule
    
    DUSt3R & \withoutmod & \withoutmod & \withoutmod & \withoutmod & \withoutmod & \firstcell{28.4} & \firstcell{75.9} & \firstcell{64.6} & \firstcell{27.2} \\
    \midrule
    {\textbf{\ours{}}} & \withoutmod & \withoutmod & \withoutmod & \withoutmod & \withoutmod 
    & \firstcell{29.8} & \firstcell{75.1} & \firstcell{66.5} & \firstcell{30.4} \\
     & \withmod & \withoutmod & \withoutmod & \withoutmod & \withoutmod & \onediffcell{60.7}{30.9} & \onediffcell{75.5}{0.4} & \onediffcell{70.3}{3.8} & \onediffcell{35.2}{4.8} \\
     & \withoutmod & \withmod & \withoutmod & \withoutmod & \withoutmod & \onediffcell{60.5}{30.7} & \onediffcell{75.8}{0.7} & \onediffcell{70.9}{4.4} & \onediffcell{37.0}{6.6} \\
     & \withmod & \withmod & \withoutmod & \withoutmod & \withoutmod & \onediffcell{89.8}{60.0} & \onediffcell{76.2}{1.1} & \onediffcell{73.7}{7.2} & \onediffcell{50.0}{19.6} \\
     & \withoutmod & \withoutmod & \withmod & \withoutmod & \withoutmod & \onediffcell{34.3}{4.5} & \onediffcell{87.9}{12.8} & \onediffcell{70.9}{4.4} & \onediffcell{35.3}{4.9} \\
     & \withoutmod & \withoutmod & \withoutmod & \withmod & \withoutmod & \onediffcell{34.3}{4.5} & \onediffcell{88.3}{13.2} & \onediffcell{71.0}{4.5} & \onediffcell{35.3}{4.9} \\
     & \withoutmod & \withoutmod & \withmod & \withmod & \withoutmod & \onediffcell{40.9}{11.0} & \onediffcell{94.9}{19.8} & \onediffcell{74.6}{8.1} & \onediffcell{46.5}{16.1} \\
     & \withoutmod & \withoutmod & \withoutmod & \withoutmod & \withmod & \onediffcell{34.9}{5.1} & \onediffcell{76.0}{0.9} & \onediffcell{86.6}{20.1} & \onediffcell{54.3}{23.9} \\
     & \withmod & \withmod & \withmod & \withmod & \withoutmod & \onediffcell{98.0}{68.1} & \onediffcell{\textbf{95.4}}{\textbf{20.3}} & \onediffcell{82.2}{15.7} & \onediffcell{71.0}{40.6} \\
     & \withmod & \withmod & \withoutmod & \withoutmod & \withmod & \onediffcell{90.6}{60.8} & \onediffcell{77.0}{1.9} & \onediffcell{91.1}{24.7} & \onediffcell{72.2}{41.8} \\
     & \withoutmod & \withoutmod & \withmod & \withmod & \withmod & \onediffcell{50.2}{20.4} & \onediffcell{95.0}{19.9} & \onediffcell{93.0}{26.5} & \onediffcell{70.5}{40.1} \\
     & \withmod & \withmod & \withmod & \withmod & \withmod & \onediffcell{\textbf{98.5}}{\textbf{68.6}} & \onediffcell{\textbf{95.4}}{\textbf{20.3}} & \onediffcell{\textbf{97.4}}{\textbf{30.9}} & \onediffcell{\textbf{90.2}}{\textbf{59.8}} \\
    \bottomrule
        \end{tabular}
    }
    \vspace{-0.3cm}
    \caption{\textbf{Impact of guiding at test time} for models trained at $224{\times}224$ resolution. We report performances on InfiniGen for DUSt3R, which cannot handle auxiliary modalities, and our single model with different sets of modalities; we show in green the absolute improvement \wrt the results without auxiliary modality. 
    }
    \vspace{-0.2cm}
    \label{tab:modality_study_infinigen}
\end{table}

\section{Multi-View Depth Estimation results}

\begin{table*}
\begin{center}
\renewcommand\arraystretch{1.2}
\setlength{\tabcolsep}{1pt} 
\scriptsize 
\small
\hspace{-3mm}
\resizebox{\textwidth}{!}{
\begin{tabular}{llccccrrrrrrrrrrrrc}
 \hline
\specialrule{1.5pt}{0.5pt}{0.5pt}
\multicolumn{2}{l}{\multirow{2}{*}{Methods}} & GT & GT & GT & Align & \multicolumn{2}{c}{KITTI} & \multicolumn{2}{c}{ScanNet} & \multicolumn{2}{c}{ETH3D} & \multicolumn{2}{c}{DTU} & \multicolumn{2}{c}{T\&T} & \multicolumn{3}{c}{Average} \\
\cline{7-19}
 && Pose & Range & Intrinsics &  & rel $\downarrow$ & $\tau \uparrow$ & rel $\downarrow$ & $\tau \uparrow$ & rel $\downarrow$ & $\tau \uparrow$ & rel $\downarrow$ & $\tau \uparrow$ & rel$\downarrow$ & $\tau \uparrow$ & rel$\downarrow$ & $\tau \uparrow$ & time (s)$\downarrow$ \\
\specialrule{1.5pt}{0.5pt}{0.5pt}
\multirow{2}{*}{} & COLMAP~\cite{schonberger2016structure,schonberger2016pixelwise} & $\checkmark$ & $\times$ & $\checkmark$ & $\times$ &{\bf 12.0}&{\bf 58.2} &{\bf 14.6}&{\bf 34.2}&{\bf 16.4}&{\bf 55.1}&{\bf 0.7}&{\bf 96.5}&{\bf 2.7}& {\bf 95.0} &{\bf 9.3} & {\bf 67.8} & $\approx$ 3 min  \\
&COLMAP Dense~\cite{schonberger2016structure,schonberger2016pixelwise} & $\checkmark$&$\times$ & $\checkmark$ & $\times$ & 26.9 & 52.7 & 38.0 & 22.5 & 89.8 & 23.2 & 20.8 & 69.3 & 25.7 & 76.4 & 40.2 & 48.8 & $\approx$ 3 min \\
\hline
\multirow{6}{*}{} & MVSNet~\cite{yao2018mvsnet} & $\checkmark$ & $\checkmark$ &$\checkmark$ & $\times$ & 22.7 & 36.1 & 24.6 & 20.4 & 35.4 & 31.4 & (1.8) & $(86.0)$ & 8.3 & 73.0 & 18.6 & 49.4 & 0.07 \\
& MVSNet Inv. Depth~\cite{yao2018mvsnet} & $\checkmark$ & $\checkmark$ &$\checkmark$ & $\times$ & 18.6 & 30.7 & 22.7 & 20.9 & 21.6 & 35.6 & (1.8) & $(86.7)$ & 6.5 & 74.6 & 14.2 & 49.7 & 0.32\\
& Vis-MVSNet~\cite{zhang2020visibility} & $\checkmark$ & $\checkmark$ & $\checkmark$ & $\times$ &{\bf 9.5}&{\bf 55.4}& \bf 8.9 & \bf 33.5 &{\bf 10.8}&{\bf 43.3}&{ (1.8)} &{ (87.4)} &{\bf 4.1}&{\bf 87.2} &{\bf 7.0} &{\bf 61.4} & 0.70 \\
& MVS2D ScanNet~\cite{yang2022mvs2d} & $\checkmark$ & $\checkmark$ & $\checkmark$ & $\times$ & 21.2 & 8.7 & (27.2) & (5.3) & 27.4 & 4.8 & 17.2 & 9.8 & 29.2 & 4.4 & 24.4 & 6.6 & \textbf{0.04} \\
& MVS2D DTU~\cite{yang2022mvs2d} & $\checkmark$ & $\checkmark$& $\checkmark$ & $\times$ & 226.6 & 0.7 & 32.3 & 11.1 & 99.0 & 11.6 & (3.6) & (64.2) & 25.8 & 28.0 & 77.5 & 23.1 & 0.05  \\
& MVS-Former++ DTU~\cite{cao2024mvsformer++} & $\checkmark$ & $\checkmark$& $\checkmark$ & $\times$ & 29.2 & 15.2 & 15.2 & 21.9 & 21.4 & 32.5 & ({\bf 1.2}) & ({\bf 91.9}) & 7.6 & 71.5 & 14.9 & 46.6 & 0.05  \\

\hline

\multirow{10}{*}{} & DeMon~\cite{ummenhofer2017demon} & $\checkmark$ & $\times$ &$\checkmark$ & $\times$ & 16.7 & 13.4 & 75.0 & 0.0 & 19.0 & 16.2 & 23.7 & 11.5 & 17.6 & 18.3 & 30.4 & 11.9 & 0.08  \\
& DeepV2D KITTI~\cite{teed2020deepv2d} & $\checkmark$ & $\times$ &$\checkmark$ & $\times$ & (20.4) & (16.3) & 25.8 & 8.1 & 30.1 & 9.4 & 24.6 & 8.2 & 38.5 & 9.6 & 27.9 & 10.3  & 1.43\\
& DeepV2D ScanNet~\cite{teed2020deepv2d}\ & $\checkmark$ & $\times$ &$\checkmark$ & $\times$ & 61.9 & 5.2 & (3.8) & (60.2) & 18.7 & 28.7 & 9.2 & 27.4 & 33.5 & 38.0 & 25.4 & 31.9  & 2.15 \\
& MVSNet~\cite{yao2018mvsnet} & $\checkmark$ & $\times$ &$\checkmark$ & $\times$ & 14.0 & 35.8 & 1568.0 & 5.7 & 507.7 & 8.3 & (4429.1) & (0.1) & 118.2 & 50.7 & 1327.4 & 20.1 & 0.15 \\
& MVSNet Inv. Depth~\cite{yao2018mvsnet} & $\checkmark$ & $\times$ &$\checkmark$ & $\times$ & 29.6 & 8.1 & 65.2 & 28.5 & 60.3 & 5.8 & (28.7) & (48.9) & 51.4 & 14.6 & 47.0 & 21.2 & 0.28  \\
& Vis-MVSNet \cite{zhang2020visibility} & $\checkmark$ & $\times$ &$\checkmark$ & $\times$ & 10.3 &{\bf 54.4}& 84.9 & 15.6 & 51.5 & 17.4 & (374.2) & (1.7) & 21.1 & 65.6 & 108.4 & 31.0 &0.82 \\
& MVS2D ScanNet~\cite{yang2022mvs2d} & $\checkmark$ &$\times$ &$\checkmark$ &$\times$ & 73.4 & 0.0 & (4.5) & (54.1) & 30.7 & 14.4 & 5.0 & 57.9 & 56.4 & 11.1 & 34.0 & 27.5 & \textbf{0.05}  \\
& MVS2D DTU~\cite{yang2022mvs2d} &$\checkmark$ & $\times$ & $\checkmark$ & $\times$ & 93.3 & 0.0 & 51.5 & 1.6 & 78.0 & 0.0 & (1.6) & (92.3) & 87.5 & 0.0 & 62.4 & 18.8 & 0.06 \\
& CER-MVS~\cite{cermvs} & $\checkmark$ & $\times$ & $\checkmark$ & $\times$ & 14.3 & 32.2 & 21.1 & 24.3 & 11.7 & \bf 47.5 &  4.1 & 71.3 &  6.4 & \bf 82.1 & 11.5 & 51.5 & 5.3 \\
& Robust MVD Baseline~\cite{schroppel2022benchmark} &$\checkmark$&$\times$&$\checkmark$&$\times$ &{\bf 7.1} & 41.9 & {\bf 7.4}&{\bf 38.4}&{\bf 9.0}&{ 42.6}&{\bf 2.7}&{\bf 82.0}&{\bf 5.0}&{ 75.1}&{\bf 6.3} &{\bf 56.0} & 0.06\\

\hline

\multirow{3}{*}{} & DeMoN~\cite{ummenhofer2017demon} &$\times$&$\times$ &$\checkmark$& $\|\mathbf{t}\|$ & 15.5 & 15.2 & 12.0 & 21.0 & 17.4 & 15.4 & 21.8 & 16.6 & 13.0 & 23.2 & 16.0 & 18.3 & 0.08 \\
& DeepV2D KITTI~\cite{teed2020deepv2d} &$\times$&$\times$&$\checkmark$&med& (3.1) & (74.9) & 23.7 & 11.1 & 27.1 & 10.1 & 24.8 & 8.1 & 34.1 & 9.1 & 22.6 & 22.7 & 2.07 \\
& DeepV2D ScanNet~\cite{teed2020deepv2d} &$\times$&$\times$&$\checkmark$& med &10.0 & 36.2 &\bf{(4.4)} & (54.8) & 11.8 & 29.3 & 7.7 & 33.0 & 8.9 & 46.4 & 8.6 & 39.9 & 3.57  \\
\cmidrule{2-19}
& \duster{}$^\dagger$ 224-NoCroCo &$\times$&$\times$&$\times$&med& 15.14&21.16 & 7.54&40.00 & 9.51&40.07 & 3.56&62.83 & 11.12&37.90 & 9.37 & 40.39 & \textbf{0.05} \\

 & \duster{}$^\dagger$224~\cite{wang2024dust3r} & $\times$ & $\times$ & $\times$ &              med & 15.39 & 26.69 &  (5.86) & (50.84) & 4.71 & 61.74 & 2.76 & 77.32 & 5.54 & 56.38 & 6.85 & 54.59 & \textbf{0.05} \tabularnewline
& \duster{}(repr.) 224~\cite{wang2024dust3r} & $\times$ & $\times$ & $\times$ &              med & 9.2 & 32.9 & (4.2) & (58.2) & 4.7 & 61.9 & 2.8 & 77.3 & 5.5 & 56.5 & 5.27 & 57.35 & \textbf{0.05}\tabularnewline

 & \ours{} 224 & $\times$ & $\times$ & $\times$ &               med & 7.0 & 39.7 & (4.2) & (58.2) & 4.5 & 62.5 & 2.9 & 75 & 5.4 & 57.4 & 4.80 & 58.56 & \textbf{0.05}\\
 & \ours{} 224 w/ RT & $\checkmark$ & $\times$ & $\times$  &      med & 7.0 & 39.5 & (4.2) & (58.7) & 4.4 & 63 & 2.9 & 75 & 5.2 & 58.8 & 4.74 & 59.00 & \textbf{0.05}\\
 & \ours{} 224 w/ K  & $\times$ & $\times$ & $\checkmark$ &med & 6.4 & 45 & (4.2) & (57.7) & 4.5 & 63.3 & 2.5 & 77.4 & 5.5 & 55.3 & 4.62 & 59.74 & \textbf{0.05}\\
 & \ours{} 224 w/ K+RT & $\checkmark$ & $\times$ & $\checkmark$ &   med & 6.4 & 44.6 & (4.1) & (58.1) & 4.5 & 63.2 & 2.3 & 80.8 & 5.2 & 57.6 & 4.50 & 60.86 & \textbf{0.05}\\

 & \duster{}$^\dagger$ 512~\cite{wang2024dust3r} & $\times$ & $\times$ & $\times$ &              med & 9.11 & 39.49 &  (4.93) & (60.20) & 2.91 & 76.91 & 3.52 & 69.33 & 3.17 & 76.68 & 4.73 & 64.52 & 0.13\tabularnewline
 & \duster{}(repr.) 512~\cite{wang2024dust3r} & $\times$ & $\times$ & $\times$ &              med & 5.4 & \bf 49.5 & \bf (3.1) & \bf (71.8) & 3.0 & 76 & 3.9 & 68.6 & 3.3 & 75.1 & 3.73 & 68.19 & 0.13\tabularnewline

 & \ours{} 512 & $\times$ & $\times$ & $\times$ &               med & 5.7 & 45.7 & (3.2) & (68.8) & 3.0 & 74.7 & 3.0 & 74.3 & 3.3 & 76.6 & 3.64 & 68.02 & 0.13\\
 & \ours{} 512 w/ RT & $\checkmark$ & $\times$ & $\times$  &      med & 5.7 & 45.8 & (3.2) & (69.7) & 2.9 & 75.6 & 3.3 & 71.6 & 3.2 & 77.9 & 3.66 & 68.12 & 0.13\\
 & \ours{} 512 w/ K & $\times$ & $\times$ & $\checkmark$ &  med & \bf 5.3 & 48.3 & (\textbf{3.1}) & (70.8) & 2.9 & 76 & 1.6 & 89.9 & 3.2 & 77.3 & 3.22 & 72.46 & 0.13\\
 & \ours{} 512 w/ K+RT & $\checkmark$ & $\times$ & $\checkmark$ &   med & \textbf{5.3} & {48.7} & (\textbf{3.1}) & ({71.4}) & \textbf{2.8} & \textbf{77.1} & \textbf{1.5} & \textbf{91.1} & \textbf{3.2} & \textbf{78.2} & \textbf{3.18} & \textbf{73.3} & 0.13\\

\specialrule{1.5pt}{0.5pt}{0.5pt}
\end{tabular}}
\vspace{-0.3cm}
\normalsize
\caption{
\textbf{Multi-view depth evaluation:} \ours{}, when using both pose and intrinsics, outperforms \duster{} as well as most other approaches, including both classical methods and learning-based techniques that utilize poses and depth ranges.
\textit{\duster{}}$^\dagger$ refers to the results reported in the original \duster{} paper, while `\duster{} (repr.)' denotes our re-implementation.
}
\label{tab:mvd_supp}
\end{center}
\end{table*}

 In Section 4.2 of main paper (multi-view depth evaluation), we provide a subset of all comparisons to the state of the art for the sake of space.
The full table can be found in Table~\ref{tab:mvd_supp}.
There, we present the full table of \ours{} compared to classical approaches like COLMAP~\cite{schonberger2016pixelwise}, and other learning-based approaches on multi-view depth estimation.
We evaluate the performance on KITTI~\cite{geiger2013vision}, ScanNet~\cite{dai2017scannet}, ETH3D~\cite{schops2017multi}, DTU~\cite{aanaes2016large}, Tanks and Temples~\cite{knapitsch2017tanks}, following protocol outlined in RobustMVD~\cite{schroppel2022benchmark}.
For \duster{} and \ours{} models with $224$ resolution, we naively downsize images to $224\times 224$. 
For $512$ resolution, we find the nearest aspect ratio within our training protocol and resizing such that the largest side is $512$ pixels.
We categorize the approaches into four groups: classical methods, learning-based approach utilizing camera poses and depth range, learning-based approaches with ground-truth intrinsics only, and \duster{} and \ours{}.
\ours{}, when provided with both camera pose and intrinsics significantly outperforms most of existing methods across the majority of datasets.
\ours{}-$512$ performs comparably to or slightly worse than \duster{}-$512$ but it is noteworthy that \ours{} is not consistently trained on RGB images only, and operates at almost the same number of parameters and compute.

\begin{figure*}[!htbp]
    \centering
    \includegraphics[width=\linewidth,trim=0 50 100 0,clip]{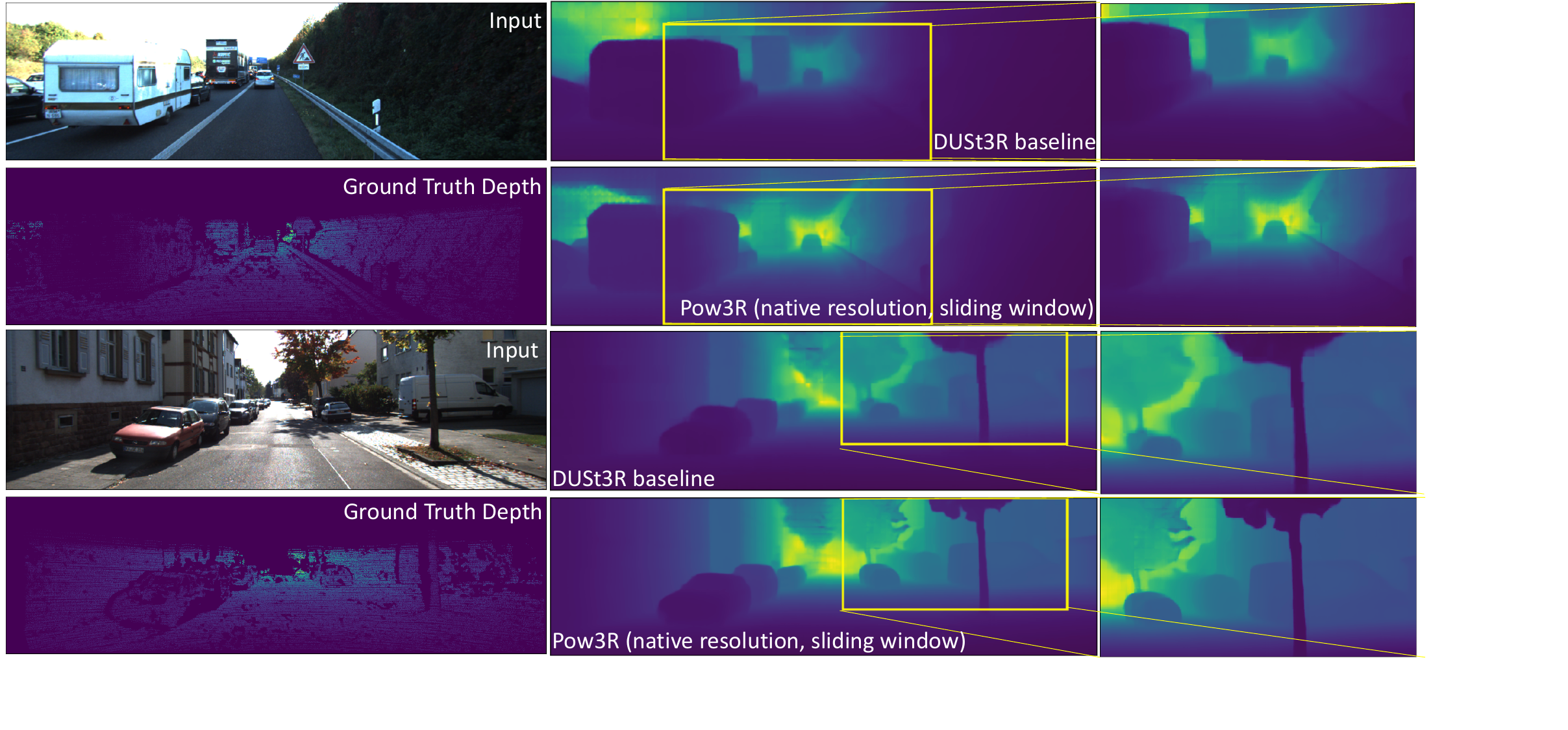} \vspace{-0.4cm}
    \caption{\textbf{High-resolution with \ours{}: } We adopt the asymmetric sliding approach as in the Figure 4 of the main paper to increase the resolution of pointmaps of a single image. \emph{Please refer to the attached video for more detailed information.}} \vspace{-0.4cm}
    \label{fig:kitti_result}
\end{figure*}

\myparagraph{Reimplementation of Evaluation Code.}
\textit{\duster{}}$^\dagger$ in Table~\ref{tab:mvd_supp} refers to the results reported in the original \duster{} paper, while `\duster{} (repr.)' denotes our re-implementation.
After observing that our re-implementation with the official code and checkpoint reaches better performance than the ones published, we have communicated with the authors to check for any issue. Authors have confirmed the presence of a bug in their internal codebase that was the cause of performance degradation.

\section{High-resolution processing with \ours{}}

\myparagraph{Overall.} 
Providing camera intrinsics as auxiliary input enables us to upsample the pointmaps by sequentially processing crops in a sliding window scheme. This is feasible since we train on non-centered cropped images along with their camera intrinsics.
As explained in Section 3.1 of the main paper, we densify the camera intrinsics as rays, and feed them in the encoder.
This allows us to deal with arbitrary aspect ratios and high-resolution images by processing image crops, which \duster{} is not capable of. A full resolution processing is not possible neither at test time nor at train time. This is clearly shown in Table 2 of the main paper, where naively feeding the high-resolution images to a low-resolution network degrades performance. Likewise, training on high resolution images is computationally prohibitive. Our multi-stage schemes, based on smaller crops, allow for processing full resolution images, without training in such high resolutions. 

\myparagraph{Asymmetric sliding.}
There are various ways to perform high-resolution processing.
In the monocular case where we would like to upsample pointmaps, we feed a downsampled coarse input image alongside the same high-resolution cropped image as shown in Figure 4 of the main paper or in \cref{fig:kitti_result} of this Supplementary.

\myparagraph{Coarse-to-fine strategy.}
Alternatively, we can feed two high-resolution image crops to the network, in which case we can condition the outcome based on an initial coarse pass.
Here, conditioning consists in feeding coarse depthmap (estimated during the initial coarse pass, where we simply downscale images) as auxiliary information for the two high-resolution crops.
The resulting pointmaps for each crop are scale-invariant; therefore, we solve their scale by simply computing the median scale factor in overlapping areas.
\emph{We refer to the attached video for additional details and visualizations.}

\myparagraph{KITTI.}
The KITTI dataset, with its unique resolution of $370\times1226$, and non-typical aspect ratio, presents a challenging test-case in a zero-shot settings. 
Using our coarse-to-fine approach, we can handle high-resolution images efficiently and produce detailed and accurate outputs as the Figure 4 of the main paper, as well as in Figure \ref{fig:kitti_result}.

\section{Controllability}

In Section 4.1 of the main paper, we quantitatively evaluate the controllability of \ours{} in Figure 6 of the paper.
In other words, we study what happens when the provided auxiliary information deviates too much from its true ground-truth value.

\myparagraph{Video results.}
\emph{We refer to the attached video showcasing the impact of providing auxiliary information for a given image pair from MegaDepth (validation set) in terms of global 3D reconstruction error.}
We observe that providing intrinsics and pose leads to noticeable improvements yet the largest impact is clearly attained when providing sparse depth, especially for pairs with large depths of field.

\begin{figure*}
    \centering
    \includegraphics[width=\linewidth,trim=0 500 250 0,clip]{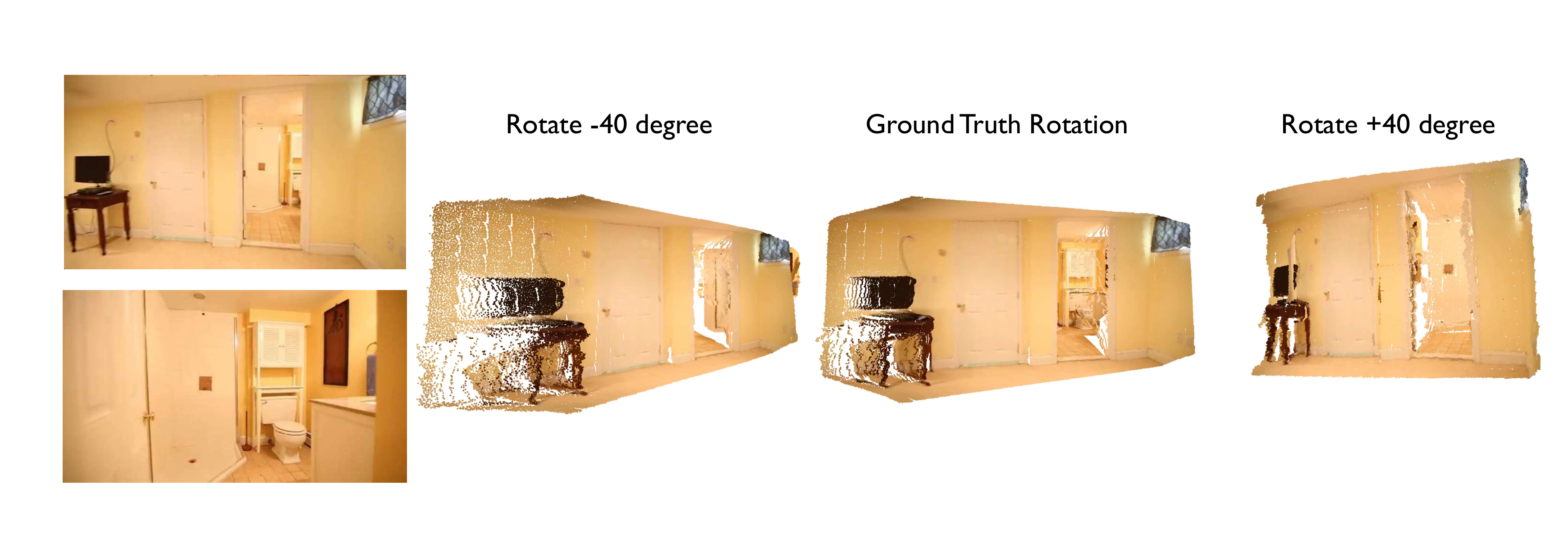} \vspace{-1.2cm}
    \caption{\textbf{Controllability test on rotation: } We provide incorrect camera rotation by -40 degree and 40 degree along y axis each in addition to the ground truth rotation, and render all of scenes from the same location. The quality of reconstruction decreases as the camera rotation deviates from the ground-truth. }
    \label{fig:rot_test}
\end{figure*}

\begin{figure*}
    \centering
    \includegraphics[width=\linewidth,trim=0 500 150 0,clip]{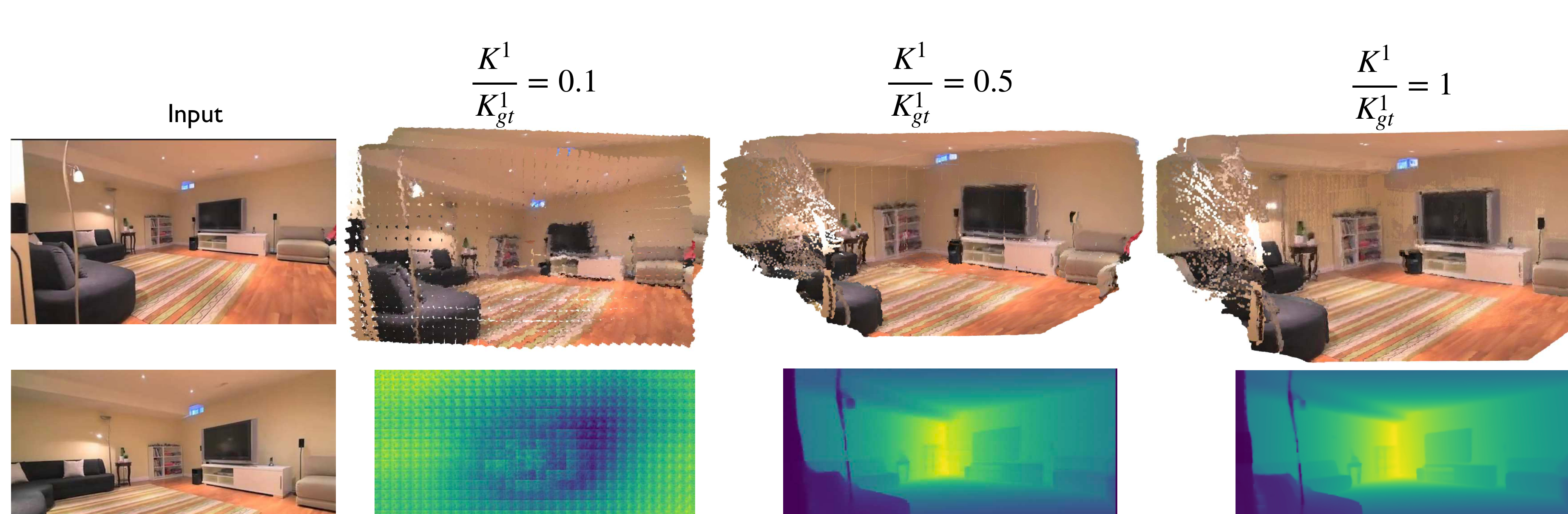}
    \vspace{-0.5cm}
    \caption{\textbf{Controllability test on focal: } We feed incorrect focal length on $K^1$, while providing the second camera with the ground-truth $K^2$. The model fails to generate accurate pointmaps for blatantly false input focals, \eg when the focal $f^1$ is set to $0.1 f^1_{gt}$. 
    The network starts to recover in this case when $f^1 \gtrsim 0.5 f^1_{gt}$.}
    \label{fig:rot_test2}
\end{figure*}

\myparagraph{Extreme cases.}
We also show qualitative results in \cref{fig:rot_test,fig:rot_test2}.
The model adheres to the guidance until it reaches a breaking point, at which point it stops functioning normally and output broken pointmaps with very low associated confidence maps, as exemplified in \cref{fig:rot_test2} for $f^1/f^1_{gt}=0.1$.


\section{Noises in the ground-truth depth annotation: NYUd - Section 4.1 of the main paper}

\begin{figure*}
    \centering
    \includegraphics[width=\linewidth,trim=0 0 440 0,clip]{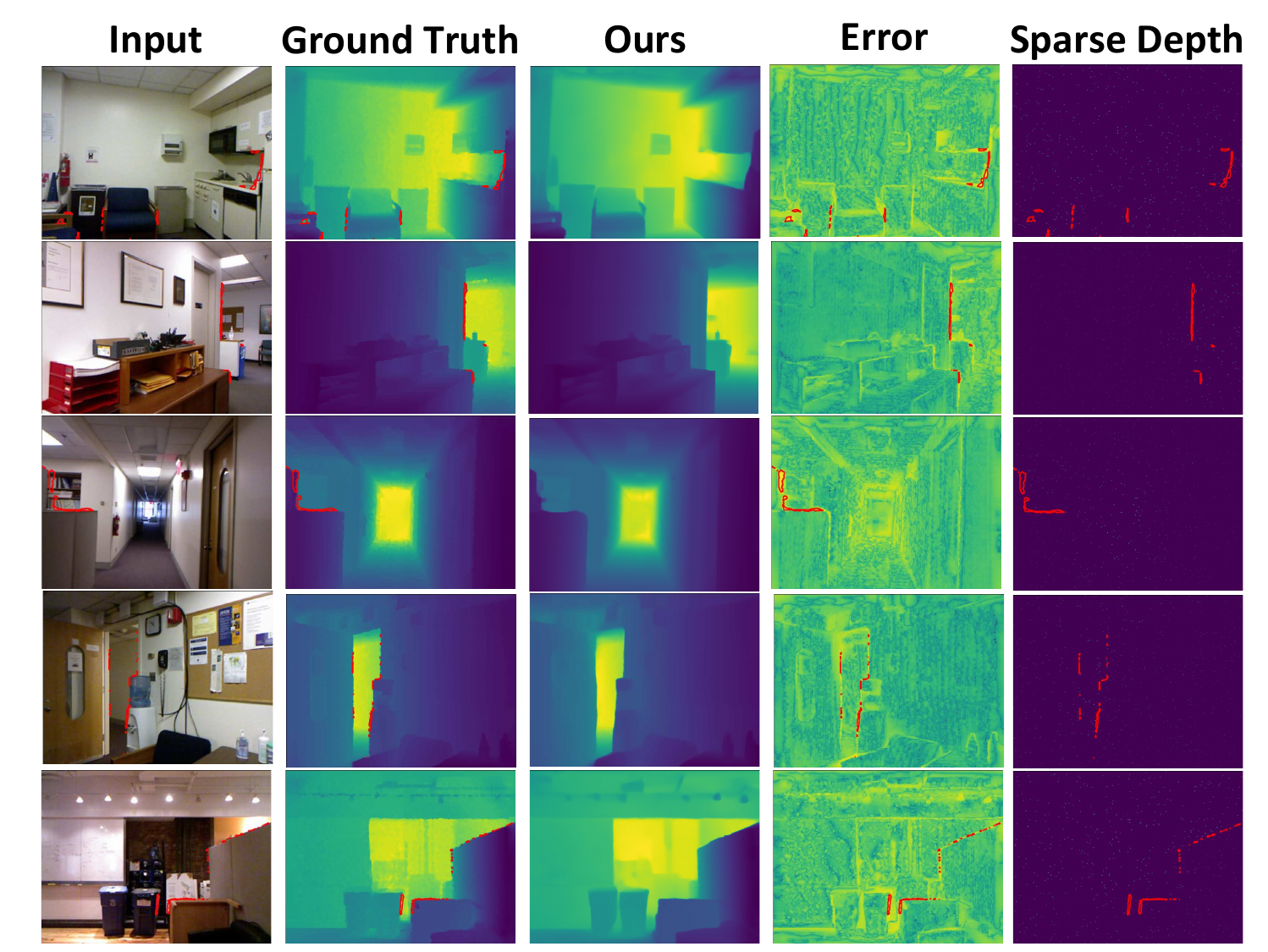}  \\[-0.3cm]
    \caption{
    \textbf{Noise in the Ground-Truth Annotations in NYUd Dataset: } Red contours in input and ground truth show areas where the error is above the threshold. Sparse depth indicates that these inaccurate annotations are not provided as the input to the network. Errors are in log scale, and \ours{} is tested on zero-shot setting.
    }
    \label{fig:nyu_errors}
\end{figure*}

In Figure \ref{fig:nyu_errors}, we highlight the erroneous ground-truth annotations present in the NYUd~\cite{SilbermanHKF12} dataset.
Specifically, the red contours in the visualization indicate regions where the discrepancy between the ground-truth and the predicted depth values exceeds a defined threshold. 
These regions often correspond to areas with edges or fine-structural details, that offer surfaces tangential to the viewing ray, and are not easy to annotate accurately.
As in Figure~\ref{fig:nyu_errors}, \ours{} can inpaint the sparse depthmap consistently and produce high-quality depthmaps. Note again that NYUd dataset is not part of our training set yet
\ours{} performs better than several depth completion models including ~\cite{completionformer2023,guidenet,bpnet,yan2024tri,lrru} across varying input sparsity depth ratios, as illustrated in Figure 5 of the main paper. We posit that a significant portion of the error observed is attributable to the aforementioned  inaccuracies in the ground-truth annotations of NYUd dataset.

\section{Extended Related work}
\label{sec:extended_related}

\myparagraph{Structure-from-Motion.} 
Traditional SfM methods typically involve non-differentiable components, such as keypoint detection, matching, and incremental camera registration; however, VGGSfM~\cite{wang2023vggsfm} integrates recent advancements in deep learning to create an end-to-end trainable system.
Graph attention networks can also be leveraged~\cite{brynte2024learning} to learn SfM by processing 2D keypoints across multiple views, and computing corresponding camera poses and 3D keypoints.
MASt3R-SfM~\cite{duisterhof2024mast3r} integrates SfM pipeline within MASt3R~\cite{leroy2024mast3r}, which eliminates the need for RANSAC by employing robust local reconstructions, and conducts optimization through successive gradient descents, first using a 3D matching loss and then refinining with a 2D reprojection loss.
\ours{} differs from traditional approaches by leveraging attention between image patches along with auxiliary inputs such as camera intrinsics, extrinsics and sparse depths, to discover camera poses directly from pointmaps.

\myparagraph{MVS and 3D reconstruction.} MVS aims to reconstruct dense 3D surface through triangulation from multiple viewpoints, traditionally with hand-crafted features~\cite{furukawa2015multi, galliani2015massively, schonberger2016pixelwise}. 
Learning-based approaches have been incorporated for MVS, followed by the emergence of Neural Radiance Fields (NeRFs) and its extended works~\cite{niemeyer2020differentiable, yariv2020multiview, gu2020cascade, leroy2021volume, peng2022rethinking, yao2018mvsnet, mildenhall2021nerf, sitzmann2019srns}.
The need for camera parameters and sparse scene initialization pushed NeRF and Gaussian Splatting (GS)~\cite{kerbl20233d} based models to leverage SfM pipelines such as COLMAP~\cite{schonberger2016structure}.
The quality of these approaches depends on the accuracy of camera parameters, and the error from cameras is not often rectified during the training.
There have been attempts to update camera parameters while optimizing the 3D scene~\cite{lin2021barf, wang2021nerf, chen2023dbarf, jeong2021self, chng2022gaussian, park2023camp}; however, many of these approaches require known camera intrinsics, good initialization, and usually rely on a weakly supervised photometric loss.
Single-view based approaches~\cite{jang2021codenerf, yu2021pixelnerf,jang2024nvist,lin2023vision,sargent2023vq3d,chen2023explicit, szymanowicz2024splatter, chan2022efficient} have been explored as they are less dependent upon camera poses, but these models usually require aligned datasets or cannot resolve the scene ambiguity completely.
The \duster{} framework departs from these approaches as it aims to do unconstrained 3D reconstruction via supervised pointmap regression, without relying on camera parameters.
In \ours{}, we further develop \duster{} by allowing the network to take auxiliary inputs such as sparse depth, camera intrinsics and camera pose. Naturally incorporating existing scene and camera priors seamlessly with RGB images further improves performance, and importantly it enables full-resolution processing of images, which was not easily achievable prior to this work~\cite{winwin}.

\myparagraph{RGB-to-3D.} 
From a single image, combined with monocular depth estimators and camera intrinsics, networks can predict pixel-aligned 3d point clouds~\cite{bian2022auto, szymanowicz2024flash3d, yin2022towards, yin2021learning}.
SynSin~\cite{wiles2020synsin} does new-view synthesis by predicting depth, generating point clouds, and using the differentiable renderer to synthesize images, and it computes the camera intrinsics by temporal consistency within video frames or off-the-shelf estimator.
For multi-view settings, ~\cite{teed2020deepv2d, ummenhofer2017demon, zhou2018deeptam} have been proposed to build a differentiable SfM, but the camera intrinsics are required.

\myparagraph{Focal Estimation.}
Classical approaches rely on parallel lines~\cite{grompone2010lsd} that intersect at vanishing points for single-image calibration, and vanishing point estimations were in~\cite{bernard1983interpreting, kluger2020consac, zhai2016detecting, tong2022transformer}.
Learning-based methods~\cite{lopez2019deep, hold-geoffroy2022deep,zhai2016detecting} were introduced to regress or classify camera parameters into bins, but there were not as accurate as traditional approaches.
Recent methods combine both traditional and learning-based approaches~\cite{zhu2023tame, veicht2024geocalib}. In our case, the focal length can be directly recovered from the pointmap representation; Our contribution is orthogonal to these lines of work in the sense that \ours{} optionally incorporate sparse depth and relative pose to improve the quality of prediction, again unlike~\duster{}.

\myparagraph{Guiding 3D.} 
Several Simultaneous Localization and Mapping (SLAM) methods incorporate both RGB and RGB-D images like ORB-SLAM2~\cite{mur2017orb}.
DROID-SLAM~\cite{teed2021droid} is a neural-network based system for SLAM that process visual data from monocular, stereo and RGB-D cameras, and it can leverage stereo or RGB-D inputs at test time indifferently. These pipelines however are heavily engineered. In this work, we wish to follow the philosophy from \duster{} where a single network regresses all relevant information, optionally leveraging auxiliary information. %

\begin{figure*}
    \centering
    \includegraphics[width=0.7\linewidth,trim=0 0 630 0,clip]{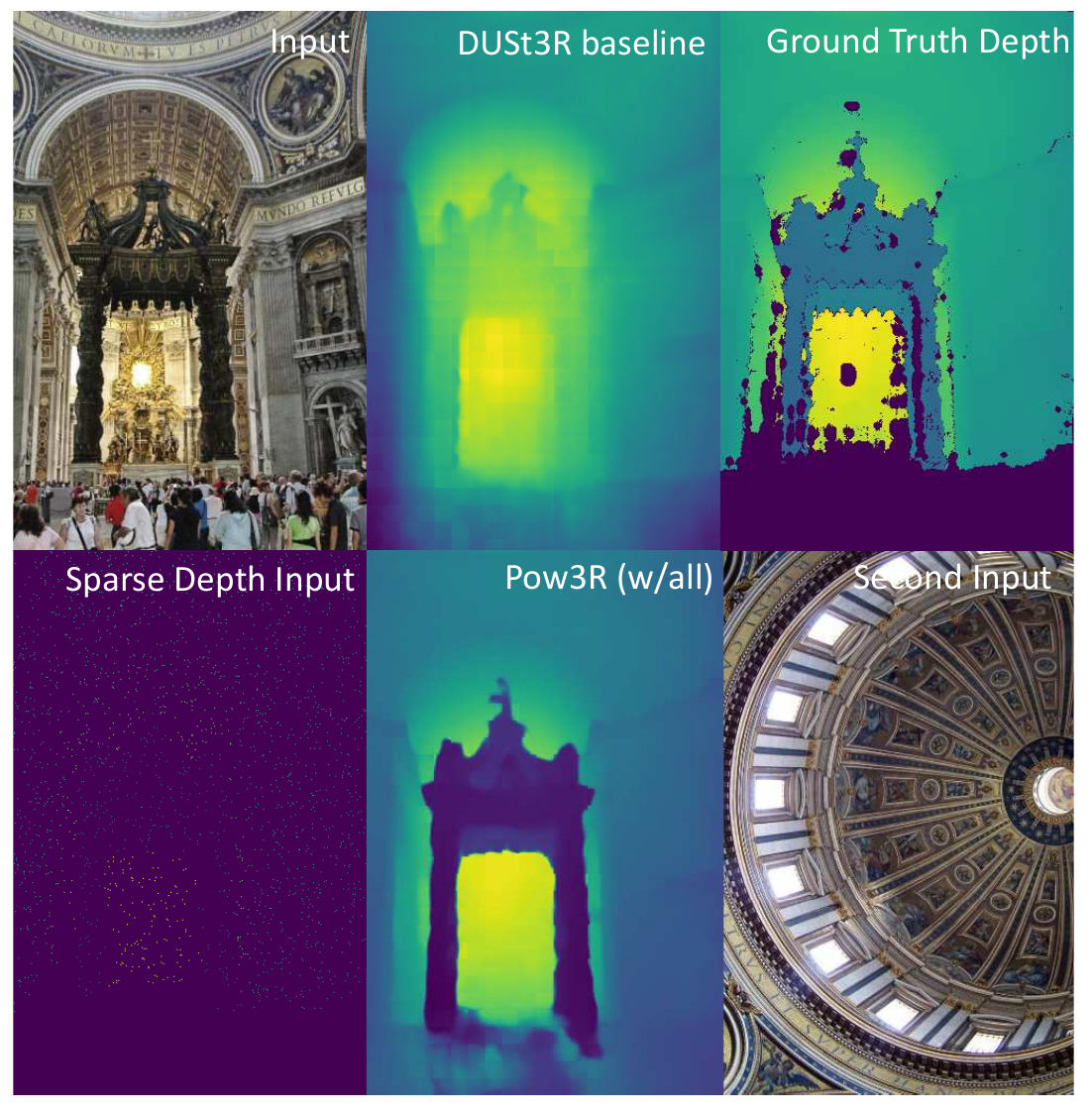} \vspace{-0.5cm}
    \caption{\textbf{Qualitative Result on depthmap.} We validate \ours{} with \duster{} on a Megadepth~\cite{li2018megadepth} indoor scene. Here, we provide camera intrinsics, pose and 2048 sparse point clouds. The result shows that \ours{} can reconstruct the gate properly and smoother than the ground point clouds. In contrast, \duster{} struggles to generate the gate properly.}
    \label{fig:qualitative_megadepth_gate}
\end{figure*}

\section{More Qualitative Results}

We showcase the impact of \ours{} when combined with auxiliary input. 
In the following, we provide examples results both in terms of depthmap and pointmap predictions. 
For each figure, the auxiliary information given to the network are the intrinsics, relative poses and $2048$ sparse depth values except RealEstate10K~\cite{realestate10K} for which no depth information is available.

\begin{figure*}
    \centering
    \includegraphics[width=\linewidth,trim=0 0 110 0,clip]{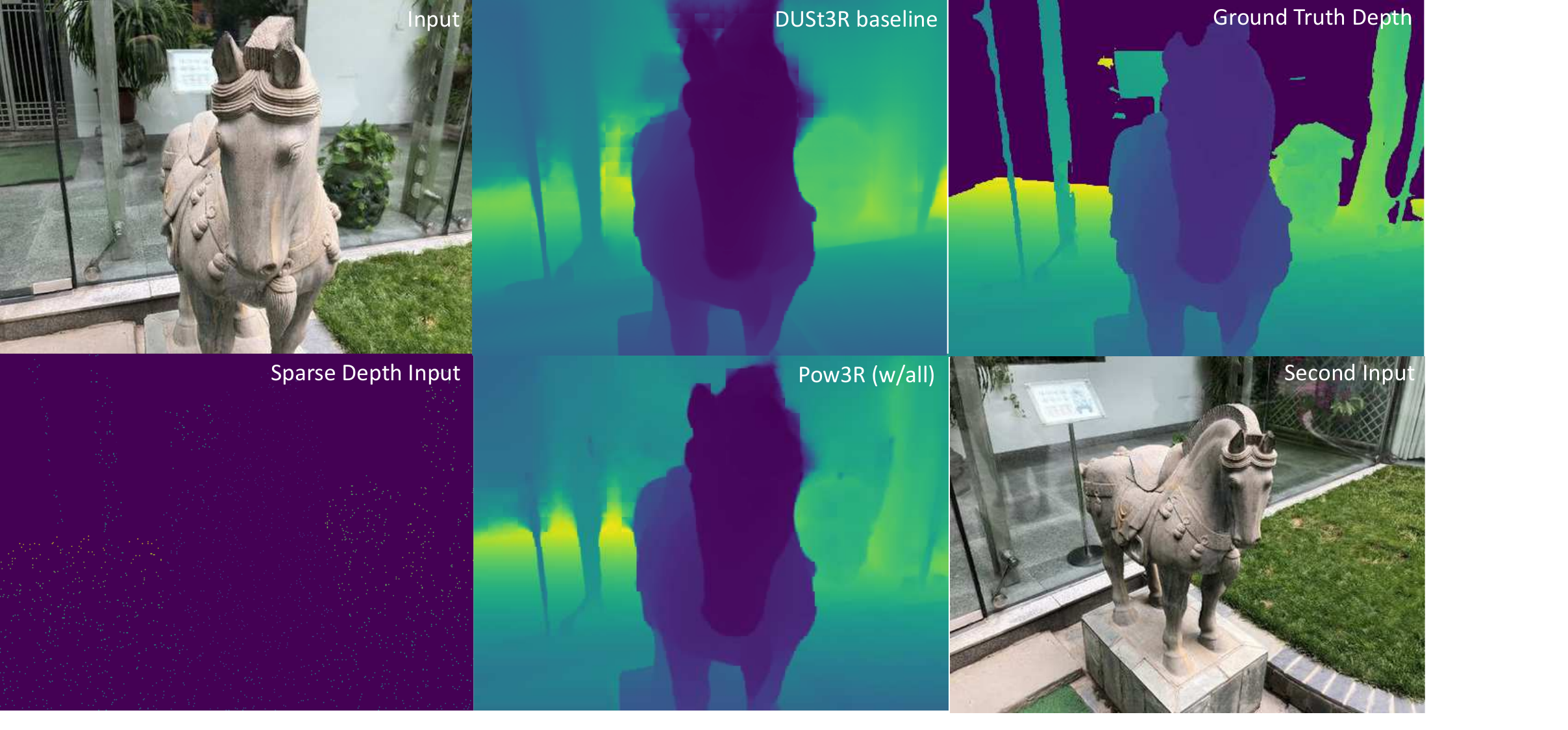} \vspace{-0.8cm}
    \caption{\textbf{Qualitative Result on depthmap.} We compare \ours{} with \duster{} on one of the BlendedMVS~\cite{yao2020blendedmvs} scenes. For \ours{}, we provide camera intrinsics, extrinsic and 2048 sparse point clouds. While \duster{} is able to capture the overall scene, it fails to correctly reconstruct the head  of the horse. In contrast, \ours{} delivers a more precise reconstruction. }
    \label{fig:qualitative_blendedmvs_statue}
\end{figure*}

\begin{figure*}
    \centering
    \includegraphics[width=0.8\linewidth,trim=0 0 580 0,clip]{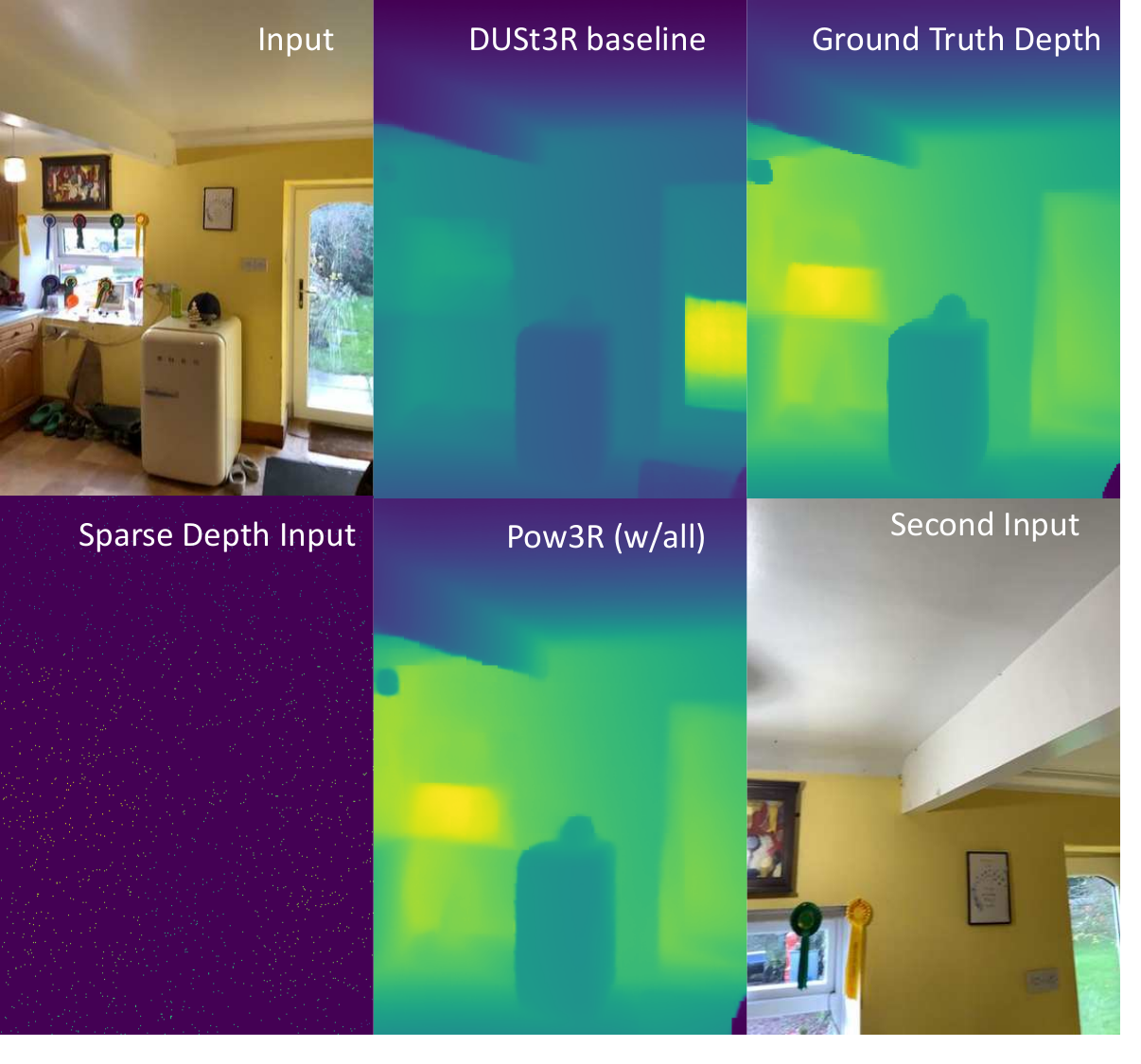} \vspace{-0.5cm}
    \caption{\textbf{Qualitative Result on depthmap.} We evaluate \ours{} against \duster{} on an indoor scene from the ARKit~\cite{baruch2021arkitscenes} dataset.%
    We feed to \ours{} camera intrinsics, pose and 2048 sparse depthmap. While \duster{} generally builds a good depthmap, but \ours{} faithfully reconstructs the glass window of the door and small objects on the fridge.}
    \label{fig:qualitative_arkit_room}
\end{figure*}

\myparagraph{Using sparse depthmaps.}
We compare depthmaps predicted by from \ours{} and \duster{} in terms of visual quality in Figures~\ref{fig:qualitative_megadepth_statue},~\ref{fig:qualitative_megadepth_gate},~\ref{fig:qualitative_blendedmvs_statue},~\ref{fig:qualitative_arkit_room}.
Results for \ours{} are consistently better, with much less failure cases than with \duster{}, which is expected given \ours{} receives additional priors.

\begin{figure*}
    \centering
    \includegraphics[width=\linewidth,trim=0 50 130 0,clip]{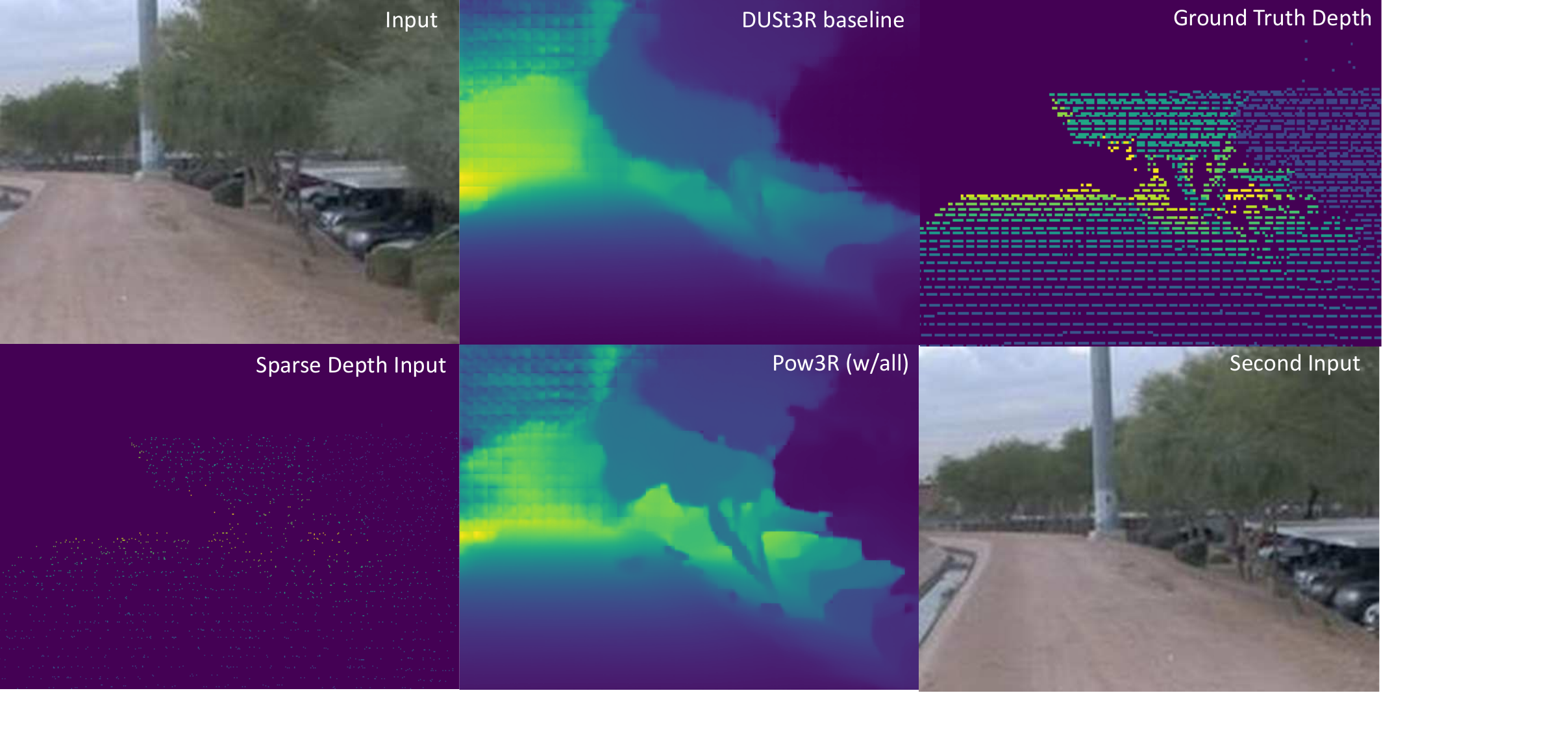} \vspace{-0.5cm}
    \caption{\textbf{Qualitative Result on depthmap.} We conduct a comparison on an outdoor scene from the Waymo~\cite{sun2020scalability} dataset. We provide to \ours{} the camera intrinsics, pose and 2048 sparse point clouds from LiDAR. While \duster{} generates a good depthmap from RGB images only, \ours{} shows better performance at capturing details of cars in the parking lot and trees.}
    \label{fig:qualitative_waymo_outside}
\end{figure*}

\begin{figure*}
    \centering
    \includegraphics[width=\linewidth]{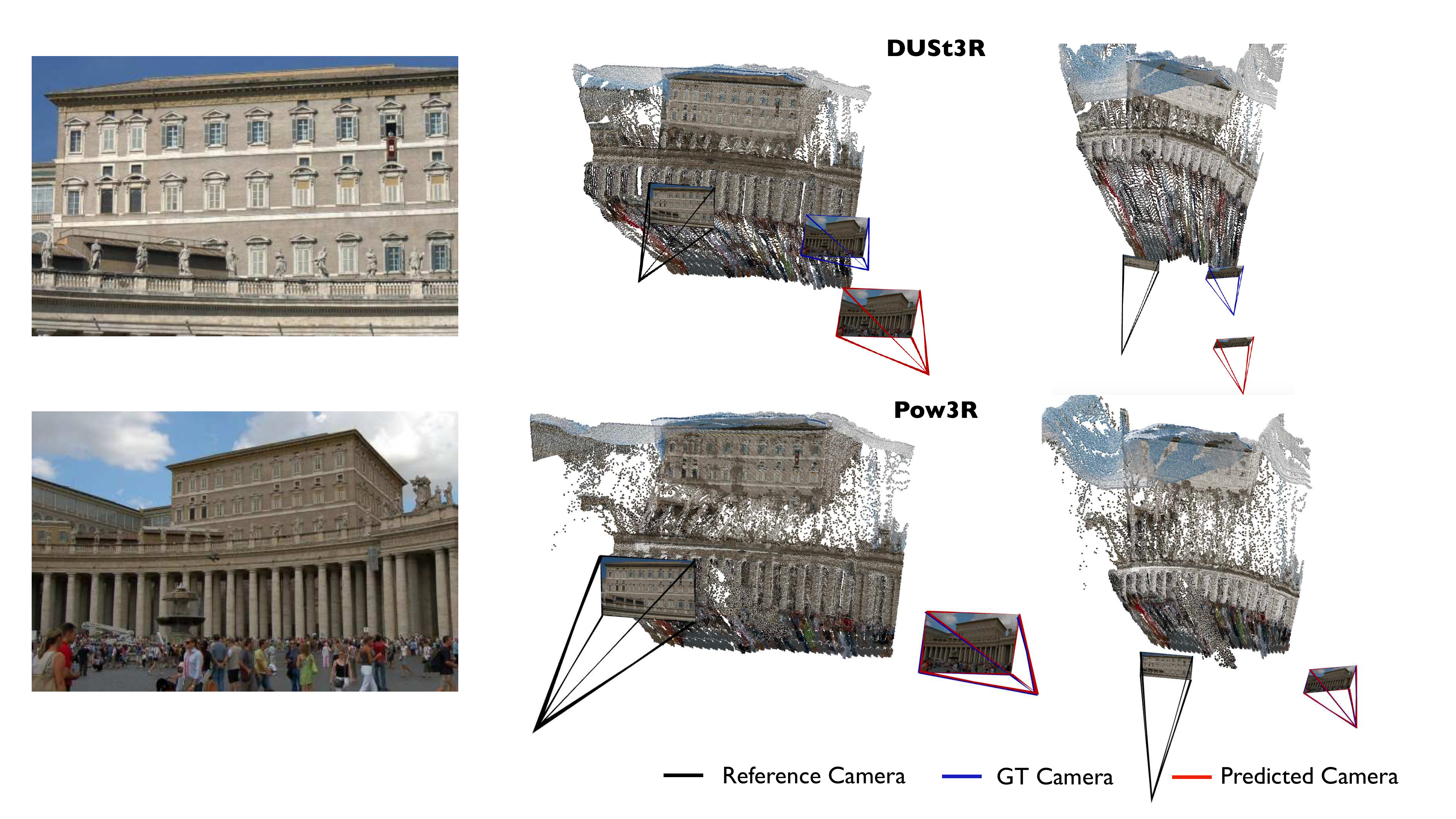}  \vspace{-0.5cm}
    \caption{\textbf{Qualitative Result on 3D reconstruction and camera estimations.} We evaluate our model on one of the Megadepth~\cite{li2018megadepth} outdoor scenes. Inputs include camera pose, intrinsics as well as 2048 sparse point clouds. While \duster{} attempts to reconstruct the scene from two extreme viewpoints, it struggles with scale ambiguity and improper camera registration. In contrast, \ours{} achieves better reconstruction as well as accurate camera registration.}
    \label{fig:qualitative_megadepth}
\end{figure*}

\begin{figure*}
    \centering
    \includegraphics[width=\linewidth]{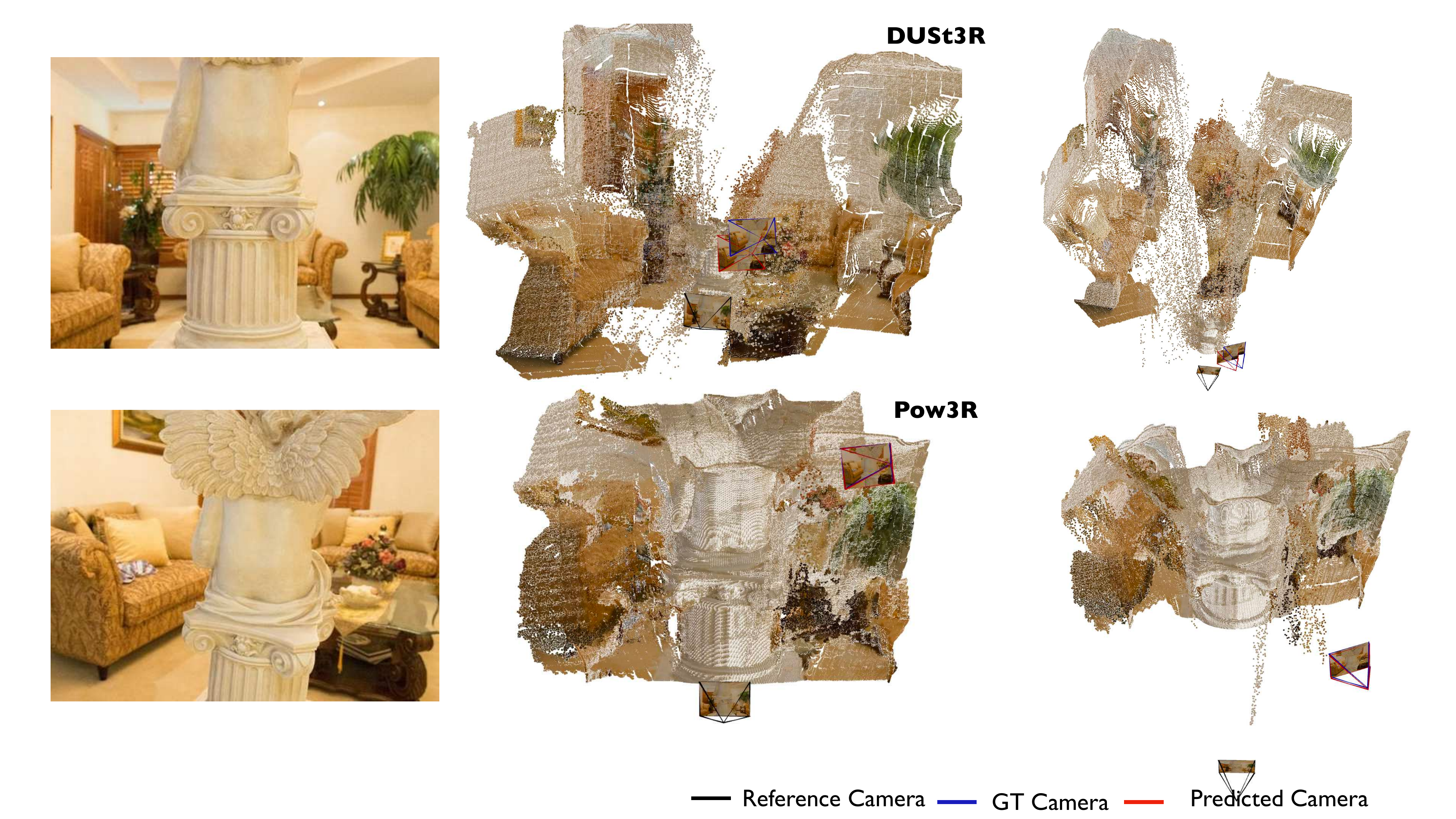} \vspace{-0.5cm}
    \caption{\textbf{Qualitative Result on 3D reconstruction and camera estimations.} We evaluate our model on one of theBlendedMVS~\cite{yao2020blendedmvs} indoor scene. We provide camera intrinsics, extrinsic and 2048 sparse depthmap. While \duster{} incorrectly predicts the depth of field and struggles with the statue, \ours{} generates the 3D scene along with cameras accurately.}
    \label{fig:qualitative_blendedmvs}
\end{figure*}

\begin{figure*}
    \centering
    \includegraphics[width=\linewidth]{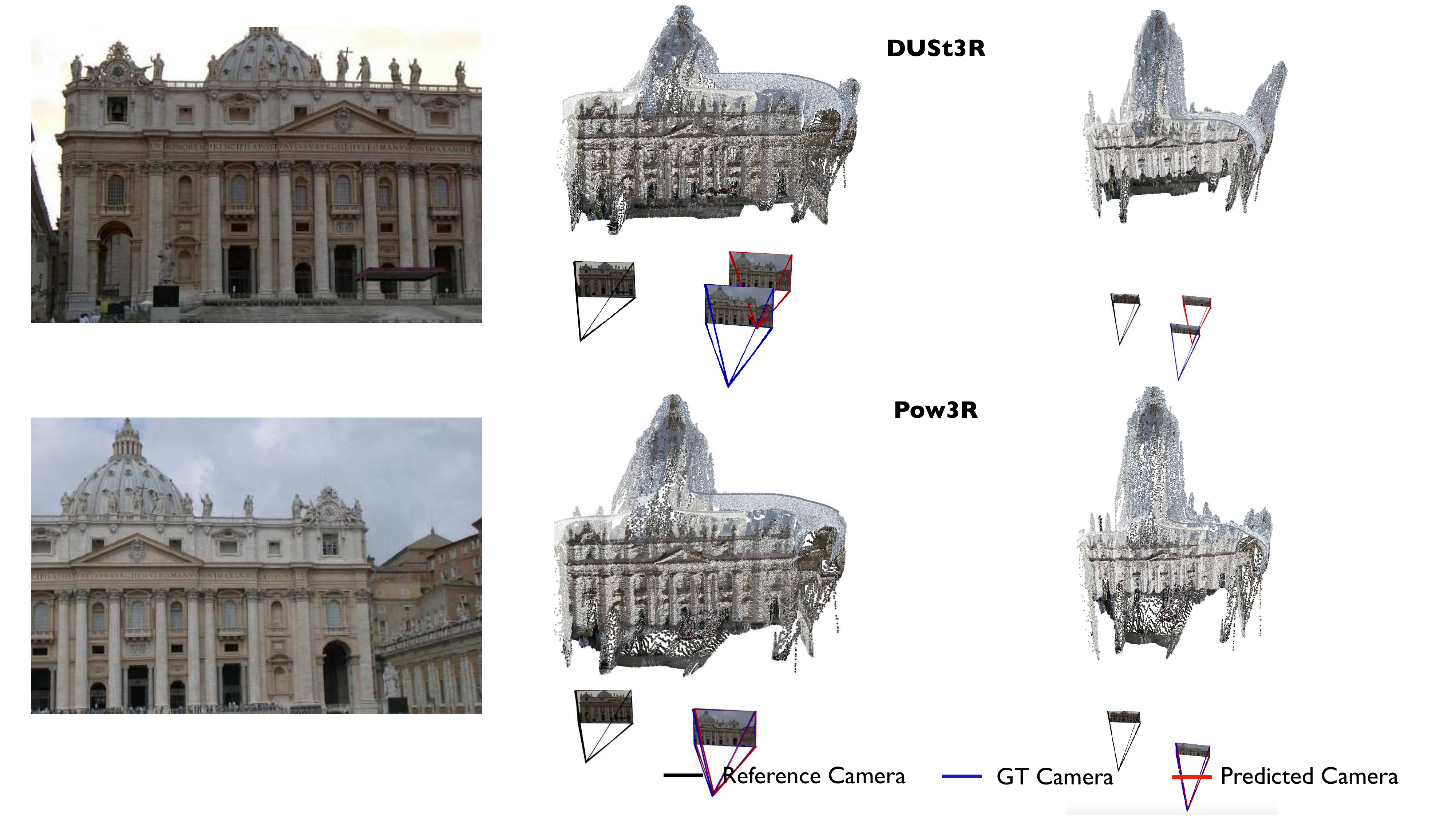} \vspace{-0.5cm}
    \caption{\textbf{Qualitative Result on 3D reconstruction and camera estimation} on an outdoor scene from the Megadepth~\cite{li2018megadepth} dataset. We feed camera intrinsics, pose and 2048 sparse depths. \ours{} excels at reconstructing the depth of field and the camera locations, contrary to \duster{}.}
    \label{fig:qualitative_megadepth2}
\end{figure*}

\begin{figure*}
    \centering
    \includegraphics[width=\linewidth]{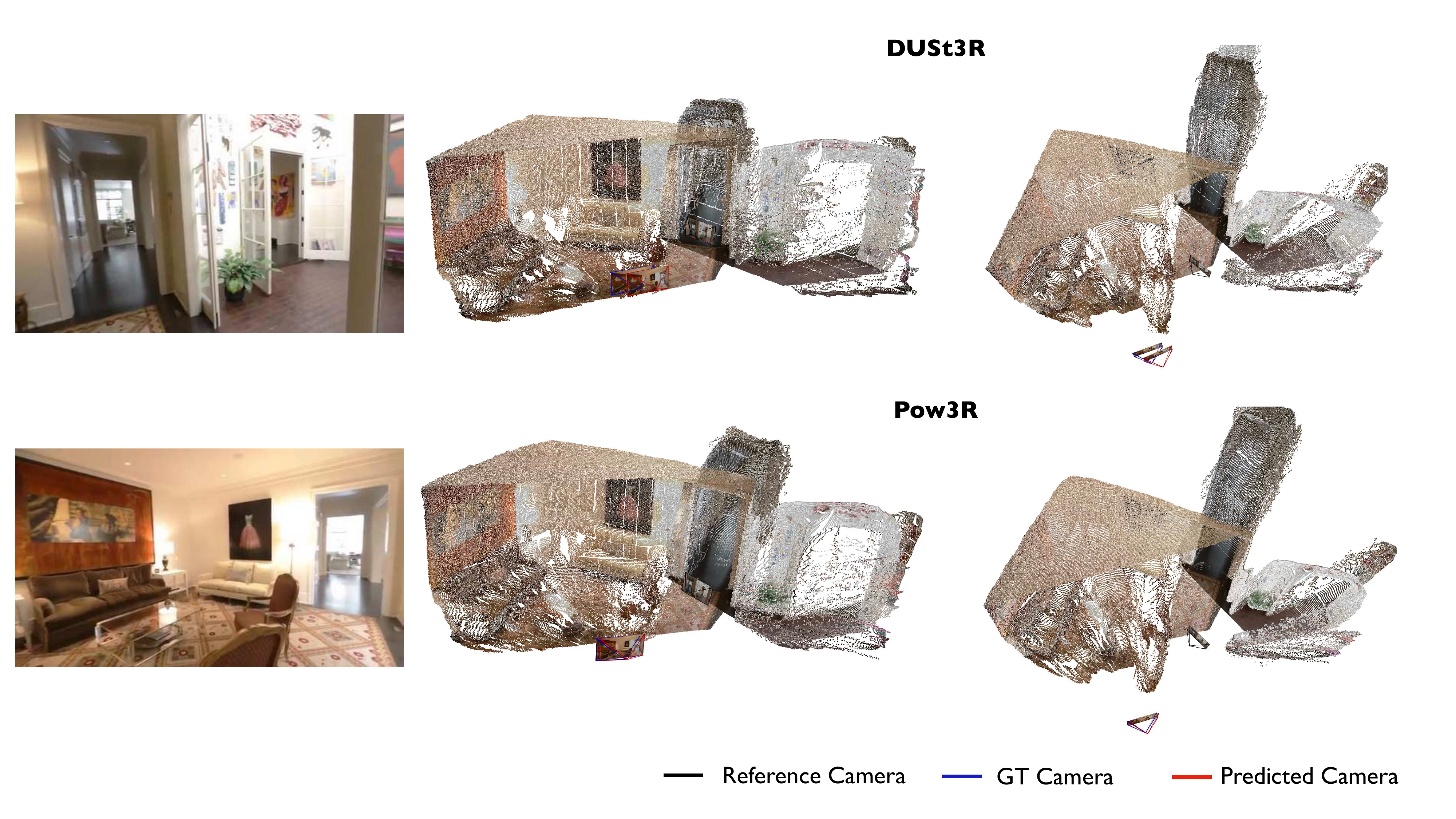} \vspace{-0.8cm}
    \caption{\textbf{Qualitative Result on 3D reconstruction and camera estimation} on an indoor scene from RealEstate10K~\cite{realestate10K} dataset. In this evaluation, we only provide the camera intrinsics and extrinsic, as RealEstate10K dataset does not have point clouds or depthmaps. Both \ours{} and \duster{} produce faithful 3D reconstructions from two diverging viewpoints,  \ours{} demonstrates better performance at predicting camera locations than \duster{}.}
    \label{fig:qualitative_re10k}
\end{figure*}

\begin{figure*}
    \centering
    \includegraphics[width=\linewidth]{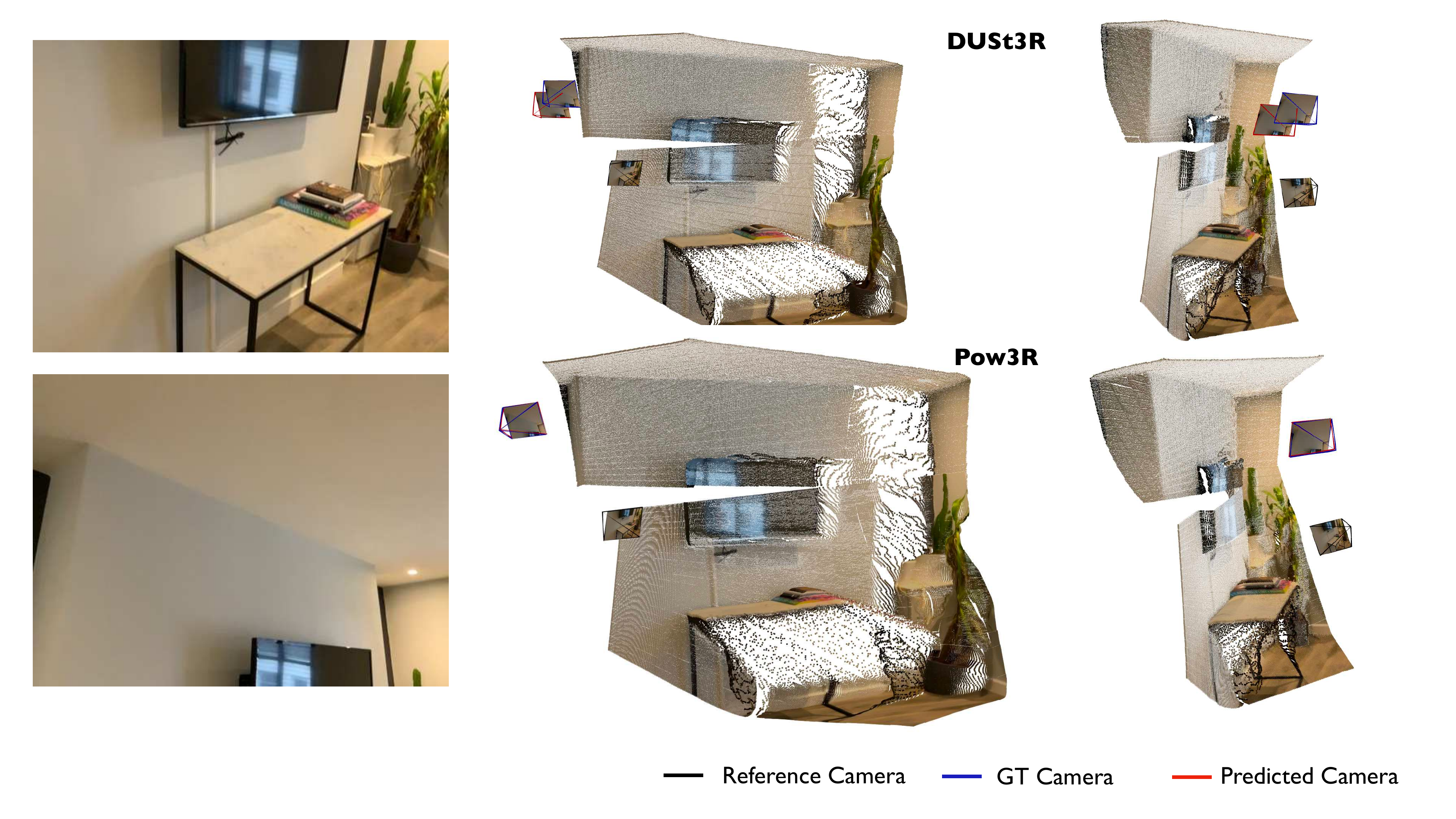} \vspace{-0.8cm}
    \caption{\textbf{Qualitative Result on 3D reconstruction and camera estimation} on an indoor scene from ARKit~\cite{baruch2021arkitscenes} dataset. We provide to \ours{} camera intrinsics, pose and 2048 sparse depth points. Both \ours{} and \duster{} generate a reasonable 3D scene from two almost non-overlapping viewpoints, \ours{} providing more accurate camera locations than \duster{}.}
    \label{fig:qualitative_arkit}
\end{figure*}

\myparagraph{Visualizing 3D pointmaps with Cameras.}
Likewise, we showcase the impact of auxiliary information against \duster{}, this time in terms of overall 3D reconstruction as well as camera locations in Figures~\ref{fig:qualitative_megadepth},\ref{fig:qualitative_blendedmvs},\ref{fig:qualitative_megadepth2},\ref{fig:qualitative_re10k},\ref{fig:qualitative_arkit}.
\ours{} reconstructs 3D scenes better than \duster{} in general, while \duster{} generates 3D scenes almost on par with \duster{} in some indoor scenes like Figures~\ref{fig:qualitative_re10k},~\ref{fig:qualitative_arkit}.
Even in these scenes, \ours{} predicts the camera location more precise than \duster{}, which demonstrates that \ours{} performs better with more auxiliary inputs.

\end{document}